\documentclass[10pt]{article}
\usepackage[utf8]{inputenc}
\usepackage[T1]{fontenc}
\usepackage{amsmath, amssymb, amsfonts}
\usepackage[margin=1in]{geometry}
\usepackage{graphicx}
\usepackage{xcolor}
\usepackage{booktabs}
\usepackage{float}
\usepackage{caption}
\usepackage{placeins}
\usepackage{tikz}
\usepackage{natbib}
\usepackage{xurl}
\usepackage{hyperref}
\hypersetup{
  colorlinks=true,
  linkcolor=blue,
  citecolor=blue,
  urlcolor=blue
}

\usetikzlibrary{arrows.meta, positioning, calc, shapes.geometric, shapes.misc, shadows, backgrounds, fit}

\definecolor{brandblue}{RGB}{0, 114, 189}
\definecolor{brandteal}{RGB}{0, 168, 142}
\definecolor{brandgray}{RGB}{245, 247, 249}
\definecolor{darkgray}{RGB}{50, 50, 50}
\definecolor{accentorange}{RGB}{255, 120, 0}
\definecolor{modernpurple}{RGB}{155, 89, 182}

\tikzset{
    tensor/.style={draw=black!70, thick, fill=blue!5, rectangle, minimum width=4cm,
        minimum height=0.9cm, align=center, font=\sffamily\small},
    latent/.style={tensor, fill=yellow!10, draw=orange!50!black},
    network/.style={draw, thick, rounded corners=4pt, minimum width=4cm, minimum height=1cm,
        align=center, font=\sffamily\small\bfseries},
    cdd_net/.style={network, top color=white, bottom color=orange!25, draw=orange!80!red,
        line width=1.2pt, minimum width=4.5cm},
    enc_net/.style={network, fill=green!10, draw=green!50!black},
    pred_net/.style={network, fill=purple!10, draw=purple!50!black},
    operator/.style={draw=black!70, thick, circle, fill=gray!15, minimum size=0.9cm,
        inner sep=0pt, font=\sffamily\Large},
    generator/.style={draw=gray!50, dashed, thick, fill=gray!5, rectangle, rounded corners=2pt,
        align=center, font=\scriptsize\sffamily, inner sep=4pt},
    loss/.style={draw=red!60!black, thick, fill=red!10, ellipse, minimum width=4.5cm,
        minimum height=1cm, align=center, font=\sffamily\small\bfseries},
    arrow/.style={-Stealth, thick, draw=gray!80},
    ema/.style={-Stealth, dashed, thick, draw=blue!50},
    signal/.style={-Stealth, dotted, thick, draw=gray!60},
    backprop/.style={-Stealth, thick, draw=red!60}
}

\tikzset{
    >={Stealth[length=3mm, width=2mm]},
    backbone/.style={draw, fill=blue!10, rectangle, minimum height=3.5cm, minimum width=2.5cm, align=center, rounded corners=3pt, thick},
    pyramid/.style={draw, fill=green!5, rectangle, minimum height=6cm, minimum width=7cm, align=center, rounded corners=3pt, thick},
    attention/.style={draw, fill=red!5, rectangle, minimum height=6cm, minimum width=3cm, align=center, rounded corners=3pt, thick},
    func/.style={draw, fill=gray!10, rectangle, minimum size=8mm, align=center, inner sep=2pt, font=\small},
    conv/.style={func, fill=yellow!20},
    up/.style={func, fill=orange!20},
    gapmlp/.style={func, fill=purple!10},
    softmax/.style={draw, fill=cyan!10, circle, minimum size=1cm, inner sep=0pt, font=\small},
    feat/.style={inner sep=1pt, font=\small},
    ops/.style={inner sep=1pt, font=\footnotesize\itshape},
    point/.style={circle, fill=black, inner sep=0pt, minimum size=3pt},
    plus/.style={circle, draw, minimum size=4mm, inner sep=0pt, font=\bfseries},
    mul/.style={circle, draw, minimum size=4mm, inner sep=0pt, font=\bfseries}
}
\title{ScaleAware-JEPA: Latent Representation for Discovery in Multiscale Physical Fields}
\author{Guang-Xing Li\thanks{Correspondence: \texttt{gxli.ai@proton.me}}}
\date{}

\begin{document}
\maketitle
\begin{abstract}
Continuous physical fields represent a large fraction of data under
scientific investigation. Their
multiscale structures are central to discovery, yet useful coordinates are
not known in advance. Standard self-supervised methods define context and
targets in fixed image coordinates, posing a predictive task misaligned with
fields organized across a continuous scale hierarchy. We introduce
ScaleAware-JEPA, a framework that constructs dense, label-free latent
coordinates for continuous scalar fields. Constrained Diffusion Decomposition
(CDD) separates each field into pixel-registered scale components and provides
the scale coordinates that define the masking geometry. The resulting JEPA
objective predicts hidden structure with a context footprint tied to the
diffusion scale of each component rather than to an arbitrary patch size. Across
MHD turbulence, interstellar molecular gas and urban nighttime-light structure,
the learned geometry maps back to coherent morphology, forming dense structural
atlases without labels or predefined segmentation rules.
By tying latent prediction to the scale hierarchy of a field, ScaleAware-JEPA
constructs latent coordinates through which complex physical patterns can be
inspected before their relevant structures have been prescribed.
Code is available at \url{https://github.com/gxli/SA-JEPA}.

\end{abstract}

\section*{Introduction}


Scientific discovery often begins when the right representation makes hidden
structure visible. Continuous physical fields have become central objects of
investigation as scientific attention shifts from isolated constituents toward
patterns, structures and collective organization. Dissipative vortex tubes in
high-Reynolds-number turbulence \citep{douady1991direct}, magnetic reconnection
layers in MHD flows \citep{biskamp2003magnetohydrodynamic}, and the anisotropic
matter distribution of the cosmic web \citep{bond1996cosmic} are all structures
whose scientific meaning lies in the organization of a field across space and
scale. Yet the representations used to analyse such fields have not kept pace.
They either compress fields into global statistics, including spectra,
probability distributions and correlation functions, discarding the spatial and
organizational structure that carries physical meaning
\citep{frisch1995turbulence,pope2000turbulent,falkovich2001particles}, or impose
hand-crafted structural definitions---Hessian filaments
\citep{aragon2007multiscale,bond2010filaments}, dendrogram clouds
\citep{rosolowsky2008decomposition} and watershed voids
\citep{platen2007cosmic}---that predetermine what can be found. A representation
that organizes the structural complexity of physical fields without compressing
it away or prescribing it in advance would provide a new instrument for
investigating systems whose relevant organizing variables are not yet known.

Learning such representations amounts to discovering useful latent
coordinates. In classical physics, the phase space of a system is often
specified by theory: the relevant variables are known before the dynamics are
analysed. For complex systems far from equilibrium, this assumption can fail.
Kauffman framed this as a foundational difficulty: the relevant macroscopic
variables, and therefore the effective phase space itself, may not be
specifiable in advance \citep{kauffman2023thirdtransition}. Existing
constructions such as Takens delay embeddings \citep{takens1981detecting} and
Koopman operator representations \citep{mezic2005spectral} provide principled coordinates for dynamical systems,
but their practical use requires choices of measurements, delays or observables
that may be difficult to specify for complex fields.

Self-supervised latent prediction provides a practical route to learning such
coordinates from the field itself. A complementary geometric perspective is
provided by the manifold hypothesis: physical configurations may occupy
structured subsets of a much larger ambient space
\citep{bengio2013representation,fefferman2016testing}. Joint Embedding
Predictive Architectures \citep[JEPAs;][]{lecun2022path,assran2023ijepa} train a
model to predict the latent representation of hidden regions from visible
context rather than reconstructing pixels. The strength of this principle is
illustrated by recent work such as LeWorldModel
\citep{maes2026leworldmodel}, which shows that JEPA-style world models
can learn compact latent dynamics directly from raw pixels using
next-embedding prediction. The aim here is not to improve representation
learning for natural images, but to construct latent coordinates through which
physical systems become inspectable.

Physical fields are intrinsically multiscale, and this poses a fundamental
challenge for standard self-supervised learning. Richardson's classical picture
of turbulence \citep{richardson1922weather,kolmogorov1941local,frisch1995turbulence}
expresses a broader organizing principle: large-scale structure sets the
environment in which smaller structures form, while small structures trace and
modify the larger organization. This coupling appears across physical domains,
from clouds and filaments to spiral arms and molecular complexes to extended
urban networks. Standard masking strategies, inherited from vision models, hide
fixed-size patches and pose prediction at a single image scale. For physical
fields, however, the context needed to predict a hidden region is distributed
across the scale hierarchy. A framework for physical fields must therefore make
that hierarchy part of the predictive task itself.

ScaleAware-JEPA instantiates this principle through two coupled design choices,
both grounded in Constrained Diffusion Decomposition
\citep[CDD;][]{li2022constrained}. First, the field is decomposed into
pixel-registered scale components before encoding (Figure~\ref{fig:framework}c),
making large-scale
correlations directly available rather than leaving them to emerge only through
the local inductive bias of a convolutional encoder. Second, the mask footprint
is tied to the diffusion scale of each component. The context--target prediction
task is thereby posed at fine, intermediate and coarse levels of the field
rather than at a single arbitrary patch size
(Figure~\ref{fig:framework}a).

\section*{Method}
\label{sec:method}

\begin{figure}[H]
\centering
\begin{minipage}[t]{0.26\textwidth}
\centering
\begin{tikzpicture}[
    node distance=1.0cm,
    auto,
    >=Stealth,
    base/.style={draw=darkgray!30, thick, rounded corners=4pt, align=center, font=\sffamily\footnotesize, fill=white},
    tensor/.style={base, fill=brandgray, minimum width=2.8cm, minimum height=0.5cm},
    cdd_net/.style={base, fill=brandblue!10, draw=brandblue!50},
    enc_net/.style={base, fill=brandteal!10, draw=brandteal!50, minimum width=1.6cm},
    pred_net/.style={base, fill=accentorange!10, draw=accentorange!50, minimum width=1.6cm},
    latent/.style={base, fill=white, draw=darkgray!60, dashed, font=\sffamily\tiny},
    loss_box/.style={base, fill=darkgray, text=white, font=\sffamily\bfseries, minimum width=1.4cm, font=\sffamily\tiny},
    generator/.style={base, fill=modernpurple!10, draw=modernpurple!50},
    arrow/.style={->, thick, color=darkgray!80},
    ema_line/.style={->, thick, color=brandblue, dashed, line width=1pt},
    back_line/.style={->, thick, color=red!60, line width=0.8pt}
]

    \node (raw) [tensor] {\textbf{Input $x$} \\ \tiny scalar field};
    \node (cdd) [cdd_net, below=0.35cm of raw] {CDD decomp.};

    \coordinate (split) at ($(cdd.south) + (0, -0.2cm)$);
    \coordinate (left_axis)  at ($(split) + (-2.0cm, 0)$);
    \coordinate (right_axis) at ($(split) + ( 2.0cm, 0)$);

    \node (mask_box) [generator] at ($(right_axis) + (0, -0.35cm)$) {Masking};
    \node (context_enc) [enc_net, below=0.4cm of mask_box] {Context ($f_\theta$)};
    \node (predict) [pred_net, below=0.4cm of context_enc] {Predictor ($g_\theta$)};
    \node (pred_lat) [latent, below=0.4cm of predict] {$\hat{z}_t$};

    \node (target_enc) [enc_net] at (left_axis |- context_enc) {Target ($f_{\bar{\theta}}$)};
    \node (target_lat) [latent] at (left_axis |- pred_lat) {$z_t$};

    \path (target_lat.south) -- (pred_lat.south) coordinate[midway] (mid_bot);
    \node (loss) [loss_box, below=0.55cm of mid_bot] {Loss};

    \draw [arrow] (raw) -- (cdd);
    \draw [thick, draw=darkgray!20] (cdd.south) -- (split);

    \draw [arrow, rounded corners=6pt] (split) -| (target_enc.north);
    \draw [arrow] (target_enc.south) -- (target_lat.north);
    \draw [arrow, rounded corners=8pt] (target_lat.south) |- (loss.west);

    \draw [arrow, rounded corners=6pt] (split) -| (mask_box.north);
    \draw [arrow] (mask_box.south) -- node[right, font=\tiny, text=darkgray]{$x_c,\,M$} (context_enc.north);
    \draw [arrow] (context_enc.south) -- (predict.north);
    \draw [arrow] (predict.south) -- (pred_lat.north);
    \draw [arrow, rounded corners=8pt] (pred_lat.south) |- (loss.east);

    \draw [ema_line] (context_enc.west) -- node[above, font=\tiny]{EMA ($m$)} (target_enc.east);

    \coordinate (bp_bus_x) at ($(pred_lat.east) + (1.0cm, 0)$);
    \draw [back_line, rounded corners=6pt] (loss.east) -- (loss.east -| bp_bus_x)
        -- (context_enc.east -| bp_bus_x)
        -- (context_enc.east);
    \node[font=\tiny, color=red!60, below=0.1cm of bp_bus_x] {Backprop};
    \draw [back_line, rounded corners=4pt] (predict.east -| bp_bus_x) -- (predict.east);

\end{tikzpicture}

\textbf{(a) Architecture}
\end{minipage}
\hfill
\begin{minipage}[t]{0.70\textwidth}
\centering
\hspace{0.10\textwidth}\scalebox{0.88}{
\begin{tikzpicture}[
    font=\footnotesize,
    >=Latex,
    box/.style={
        draw, thick, rounded corners=2pt,
        align=center,
        minimum height=0.75cm,
        minimum width=1.45cm
    },
    cddscale/.style={
        draw, thick, rounded corners=2pt,
        align=center,
        minimum height=0.72cm,
        minimum width=1.35cm
    },
    branch/.style={
        draw, thick, rounded corners=2pt,
        align=center,
        minimum height=0.72cm,
        minimum width=1.65cm
    },
    op/.style={
        draw, thick, circle,
        align=center,
        minimum size=0.62cm
    },
    arrow/.style={->, thick}
]
\node[box] (cdd) at (0,0) {CDD\\decomp.};
\node[cddscale] (s1) at (2.5, 2.4) {$x^{(1)}$\\fine};
\node[cddscale] (s2) at (2.5, 0.8) {$x^{(2)}$\\mid};
\node[cddscale] (s3) at (2.5,-0.8) {$x^{(3)}$\\coarse};
\node[cddscale] (s4) at (2.5,-2.4) {$x^{(4)}$\\halo};
\draw[arrow] (cdd.east) -- ++(0.35,0) |- (s1.west);
\draw[arrow] (cdd.east) -- ++(0.35,0) |- (s2.west);
\draw[arrow] (cdd.east) -- ++(0.35,0) |- (s3.west);
\draw[arrow] (cdd.east) -- ++(0.35,0) |- (s4.west);
\node[branch] (e1) at (5.2, 2.4) {scale-aware\\adapter};
\node[branch] (e2) at (5.2, 0.8) {scale-aware\\adapter};
\node[branch] (e3) at (5.2,-0.8) {scale-aware\\adapter};
\node[branch] (e4) at (5.2,-2.4) {scale-aware\\adapter};
\draw[arrow] (s1) -- (e1);
\draw[arrow] (s2) -- (e2);
\draw[arrow] (s3) -- (e3);
\draw[arrow] (s4) -- (e4);
\node[op] (fuse) at (6.9,0) {$\oplus$};
\draw[arrow] (e1.east) -- ++(0.30,0) |- (fuse.north);
\draw[arrow] (e2.east) -- ++(0.30,0) |- (fuse.west);
\draw[arrow] (e3.east) -- ++(0.30,0) |- (fuse.west);
\draw[arrow] (e4.east) -- ++(0.30,0) |- (fuse.south);
\node[font=\scriptsize, align=center] at (6.9,1.35) {delayed\\scale fusion};
\node[box, minimum width=1.85cm] (convnext) at (8.9,0) {dense\\ConvNeXt};
\draw[arrow] (fuse) -- (convnext);
\draw[arrow] (convnext) -- ++(1.2,0);

\end{tikzpicture}
}

\textbf{(b) Scale-aware encoder}
\end{minipage}

\vspace{1em}

\includegraphics[width=\textwidth]{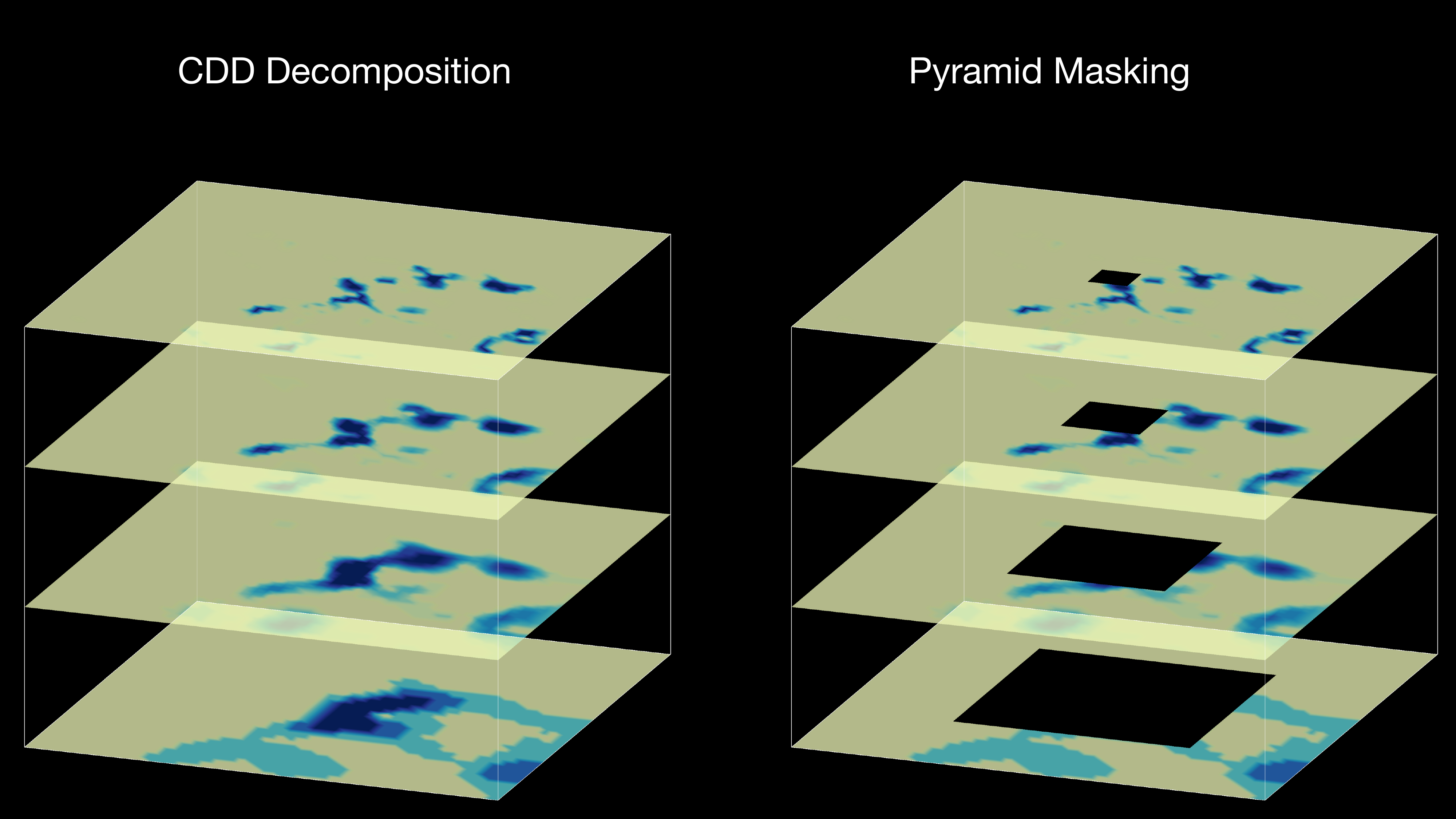}

\textbf{(c) CDD decomposition and pyramid masking}

\caption{\textbf{ScaleAware-JEPA architecture and design.}
\textbf{(a)} Architecture: the CDD frontend decomposes the raw field into scale-separated components. The context branch applies scale-aware masking and encodes the masked context; the target branch bypasses masking and uses an EMA-updated copy. A lightweight predictor maps context to target latent space. Training combines latent prediction and a weak spread regularizer. \textbf{(b)} Scale-aware ConvNeXt encoder: each CDD component is processed by a per-scale adapter, followed by top-down residual scale fusion with per-scale $1\times1$ projections and dense ConvNeXt processing to produce the latent map $z_c$. \textbf{(c)} CDD decomposition and pyramid masking.}
\label{fig:framework}
\end{figure}

ScaleAware-JEPA constructs dense, label-free latent coordinates for continuous scalar fields by reformulating joint-embedding predictive learning \citep{lecun2022path,assran2023ijepa} for multiscale continuous fields. The framework follows the core JEPA paradigm: an online branch receives a masked input and predicts the latent representation produced by an EMA target branch from the corresponding unmasked field. Prediction is evaluated strictly at hidden target locations, while a weak spread regularizer prevents representational collapse (Figure~\ref{fig:framework}).

The fundamental departure of ScaleAware-JEPA is that both the representation and the predictive task are explicitly governed by the intrinsic physical scale hierarchy of the field, rather than by arbitrary image patches. We use Constrained Diffusion Decomposition (CDD) \citep{li2022constrained} to extract a pixel-registered pyramid of continuous scale components. A scale-aware, dense ConvNeXt-style encoder \citep{liu2022convnext} maps this multiscale input to a full-resolution latent field, assigning a latent coordinate to every spatial location. Crucially, these same CDD scales dictate the masking geometry---forcing the architecture to predict hidden structure using context footprints matched to the fine, intermediate, and coarse scales of the field (Figure~\ref{fig:framework}c).

\subsection*{CDD scale coordinates}
\label{sec:cdd_coordinates}

Physical fields are observed on discrete grids but are organized across continuous physical scales. ScaleAware-JEPA uses Constrained Diffusion Decomposition \citep[CDD;][]{li2022constrained} to define scale coordinates for this organization. CDD evolves an input scalar field $I$ according to:
\begin{equation}
\frac{\partial I}{\partial t} = -\mathrm{ReLU}(-\nabla^2 I),
\label{eq:cdd}
\end{equation}
where diffusion time corresponds to a characteristic scale $\lambda \propto \sqrt{t}$ (the reported scales $[2,4,8,\dots]$ are diffusion-scale indices that parameterize the masking hierarchy consistently across experiments). Differences between diffusion states yield a pixel-registered pyramid of structurally isolated field components.

Within the ScaleAware-JEPA framework, CDD serves two coupled roles. First, instead of forcing a network to blindly infer a complex hierarchy from a single flat image, CDD presents the fine, intermediate, and coarse morphology to the encoder as physically aligned scale components. Second, it supplies the explicit coordinate system used to generate the masking task. The representation space and the predictive question are thereby locked to the exact same physical geometry.

\

We use CDD rather than a generic wavelet basis
\citep{daubechies1992ten,mallat2009wavelet} because the decomposition directly
defines the prediction task. CDD provides
localized, pixel-registered scale components well suited to defining
scale-dependent masking. Appendix~\ref{app:cdd_wavelet_control} provides a
matched wavelet control with quantitative diagnostics and a direct
decomposition comparison.

\subsection*{Scale-aware encoding and masking}
\label{sec:multiscale_masking}

For each scale level, the encoder receives three channels: the masked CDD
component $x_c^{(s)}$, a binary mask-indicator channel
$M^{(s)}\in\{0,1\}^{H\times W}$, and a scalar scale code identifying the
diffusion scale (see Figure~\ref{fig:framework}b for the scale-aware encoder
schematic). Here $M^{(s)}_{ij}=1$ denotes a masked pixel and
$M^{(s)}_{ij}=0$ a visible pixel. The mask indicator is not a learned
parameter. Adapted features are then fused from coarse to fine and
processed by a dense ConvNeXt V2-style backbone
\citep{liu2022convnext} with Global Response
Normalization \citep{woo2023convnextv2}. This produces a full-resolution latent
map while preserving the spatial correspondence required for back-mapping.
Full encoder implementation details are provided in
Appendix~\ref{app:scale_aware_encoder}.

For a component with diffusion scale $\sigma_s$ (where $\sigma_s$ is the physical scale
$\lambda_s$ expressed in pixel units), the context mask footprint is
defined relative to that scale:
\begin{equation}
b = \max(n_{\rm target},\; \operatorname{round}(\sigma_s f_{\mathrm{mask}} + B_0)), \;
B_s =
\begin{cases}
\operatorname{oddceil}(b), & b \le B_{\mathrm{cap}},\\[4pt]
\operatorname{oddfloor}(B_{\mathrm{cap}}), & b > B_{\mathrm{cap}},
\end{cases}
\end{equation}
where $n_{\rm target}=3$ is the target-patch width,
$\operatorname{oddceil}(x)$ is the smallest odd integer not smaller than
$x$, $\operatorname{oddfloor}(x)$ the largest odd integer not larger than
$x$, $f_{\mathrm{mask}}$ is the scale multiplier, $B_0$ is a fixed offset
in pixels, and $B_{\mathrm{cap}}=48$\,px (MHD, Chengdu) or
$35$\,px (NGC). Setting $f_{\mathrm{mask}}=0$ recovers a fixed-box mask;
$B_0=0$ gives a pure scale-tied pyramid mask. When the hard cap is active,
the box rounds downward to avoid exceeding $B_{\mathrm{cap}}$.
The MHD masking sweep shows that mask geometry has a direct and measurable
effect on latent-use diagnostics. Fixed-box masks and very large pyramid
footprints can reach high effective rank while the hinge ratio approaches a
plateau, indicating that further increases in occlusion no longer improve
latent usage in a controlled way. Scale-aware pyramid masking instead provides
a structured progression across scale-dependent context footprints. The
selected $1.2\times\sigma_s$ setting lies before the large-mask plateau regime
(Figure~\ref{fig:mhd_mask_sensitivity}).
The full mask-construction procedure is given in
Appendix~\ref{app:mask_construction}.

\subsection*{Learning and latent-atlas construction}
\label{sec:loss_training}

The online encoder, EMA target encoder and lightweight spatial predictor are
trained with a two-term objective:
$\mathcal{L}
= \lambda_{\mathrm{pred}}\mathcal{L}_{\mathrm{pred}}
+ \lambda_{\mathrm{spread}}\mathcal{L}_{\mathrm{spread}}$.
The dominant term $\mathcal{L}_{\mathrm{pred}}$ ($\lambda_{\mathrm{pred}}=50$) is
a mean-squared-error loss between predicted and target projected
representations, evaluated only at masked target locations. A
standard-deviation hinge spread regularizer $\mathcal{L}_{\mathrm{spread}}$
($\lambda_{\mathrm{spread}}=5$, $\tau=1$) prevents representational collapse by
penalizing latent channels whose batch standard deviation falls below the target
threshold~$\tau$.
Full loss definitions and hyperparameter defaults are provided in
Appendices~\ref{app:training_config} and~\ref{app:training_defaults}. Run sweeps are given in
Appendix~\ref{app:run_sweeps}.

At inference, masking is disabled and the EMA target encoder produces a dense
projected latent map for the full field. The latent vectors are projected with
PCA and UMAP \citep{mcinnes2018umap} for visualization and mapped back to their
spatial locations, forming a back-mappable atlas: neighborhoods, branches and
extremal regions in latent space can be traced to their original coordinates
to identify the field morphology represented by the network.

\section*{Latent Organization of MHD Turbulence}

\begin{figure}[H]
\centering
\includegraphics[width=\textwidth]{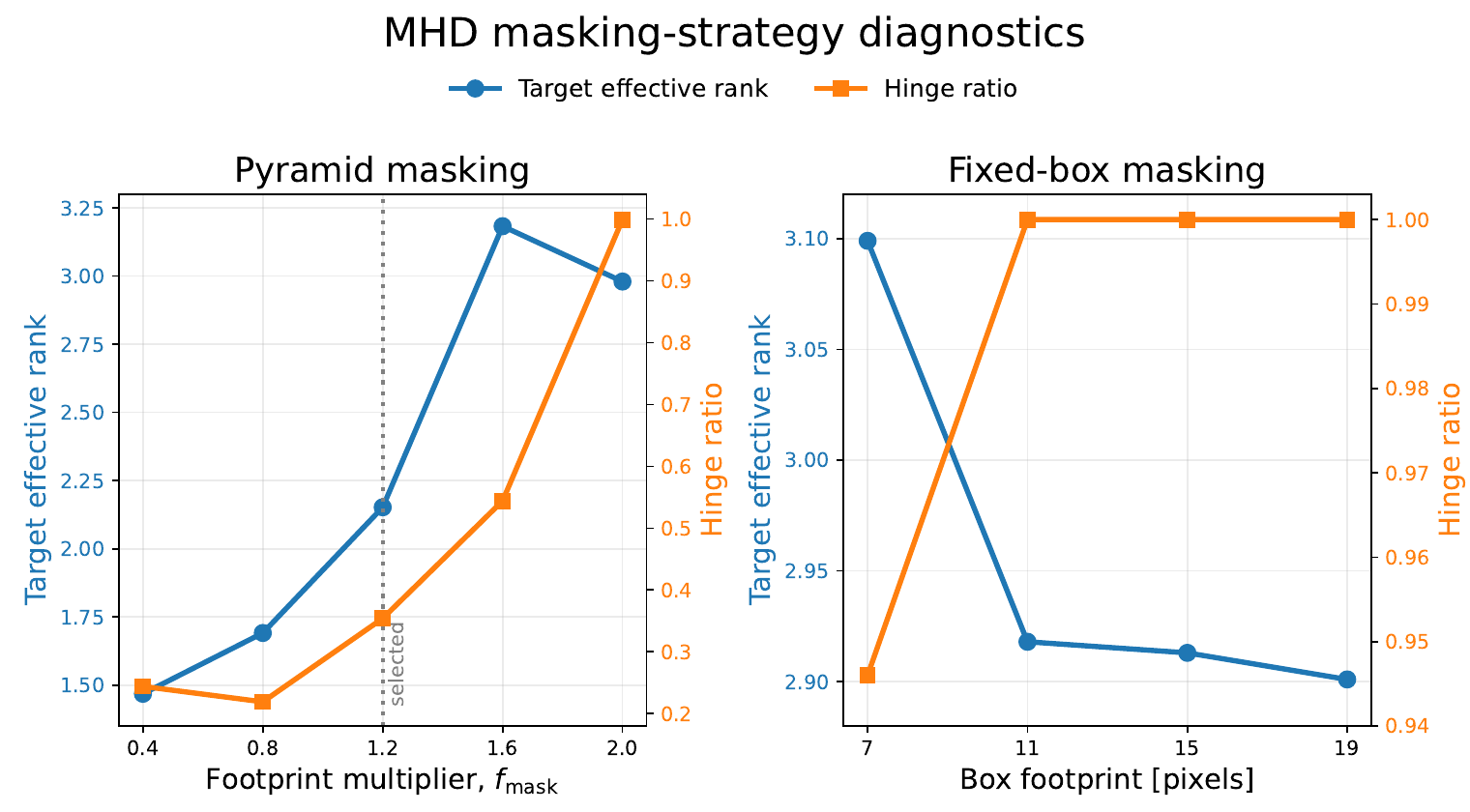}
\caption{\textbf{Masking-strategy diagnostics for the MHD sweep.}
Left: target effective rank and hinge ratio as functions of the pyramid-mask
footprint multiplier. Right: the same diagnostics for fixed-box masks.
The pyramid sweep shows a gradual increase in both diagnostics through
$1.2\times\sigma_s$, followed by a sharp rise in target effective rank at
$1.6\times\sigma_s$ and near-complete hinge saturation at
$2.0\times\sigma_s$. The selected $1.2\times\sigma_s$ setting therefore
retains an intermediate target rank and hinge response before the
large-footprint transition. For fixed-box masks, the hinge ratio is already
high at 7 px and reaches saturation by 11 px, while the target effective rank
decreases slightly and then remains nearly constant.}
\label{fig:mhd_mask_sensitivity}
\end{figure}

We evaluate ScaleAware-JEPA on a high-Reynolds-number simulation of compressible
magnetohydrodynamic (MHD) turbulence. The model receives only a two-dimensional
gas-density slice from the turbulent-cloud state-separation dataset of
\citet{Collins2012}, accessed through the CATS portal \citep{Burkhart2020}.
These fields contain diffuse material, shear interfaces, filamentary ridges,
and shock-compressed structures produced by the turbulent cascade. Previous MHD
analyses associate this density morphology spatially with magnetic, transition,
and kinetic regimes \citep{LiZhao2025mhd}. We therefore ask whether a
self-supervised representation trained on density alone can organize these
structures without labels, hand-crafted diagnostics, or auxiliary velocity and
magnetic-field inputs. Figure~\ref{fig:mhd_mask_sensitivity} summarizes the
masking-sensitivity analysis for the MHD sweep; the selected
$1.2\times\sigma_s$ pyramid-mask setting lies before the sharp
large-footprint transition and hinge-saturation regime.

\begin{figure}[H]
\centering
\includegraphics[width=\textwidth]{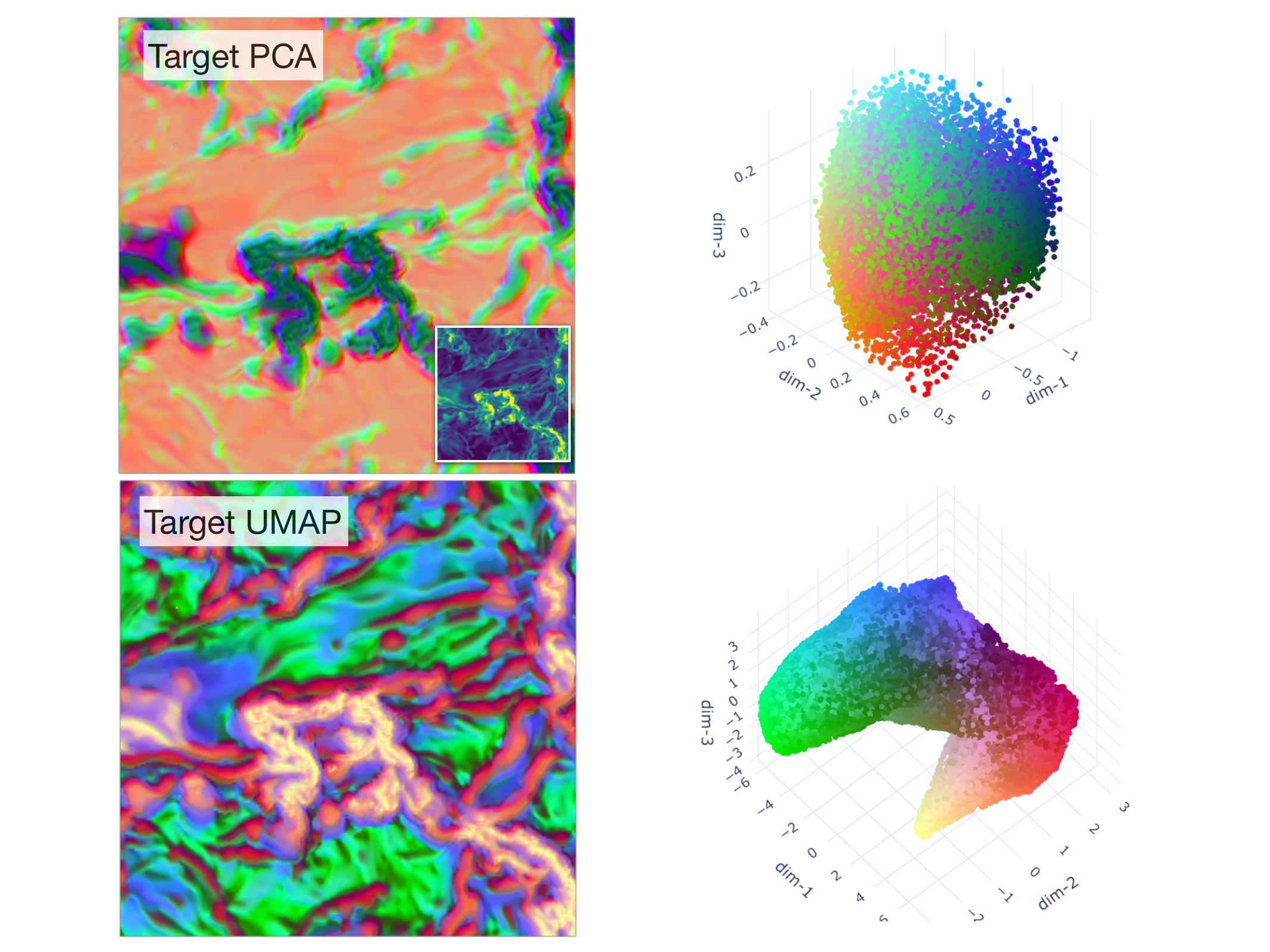}
\caption{\textbf{Dense latent topology learned for MHD turbulence.}
Target-encoder embeddings are projected to three dimensions using PCA and UMAP
and mapped back to their original spatial locations. The PCA projection provides
a linear view of the dominant latent directions, while the UMAP projection
emphasizes nonlinear neighborhood structure in the learned representation. Both
projections reveal spatially coherent organization: extended sheets, filamentary
interfaces, and compact high-contrast structures occupy distinct regions of
latent space. The inset point clouds show the corresponding three-dimensional
projected latent distributions.}
\label{fig:jepa_mhd}
\end{figure}

Figure~\ref{fig:jepa_mhd} shows dense latent coordinates from the frozen EMA
target branch. A three-dimensional PCA projection provides a linear view of the
dominant latent directions, while UMAP \citep{mcinnes2018umap} resolves local neighborhood structure.
Both reveal connected organization rather than a featureless latent cloud. We
use these projections as maps for inspecting the learned representation, not as
a unique physical phase diagram.

\begin{figure}[H]
\centering
\includegraphics[width=\textwidth]{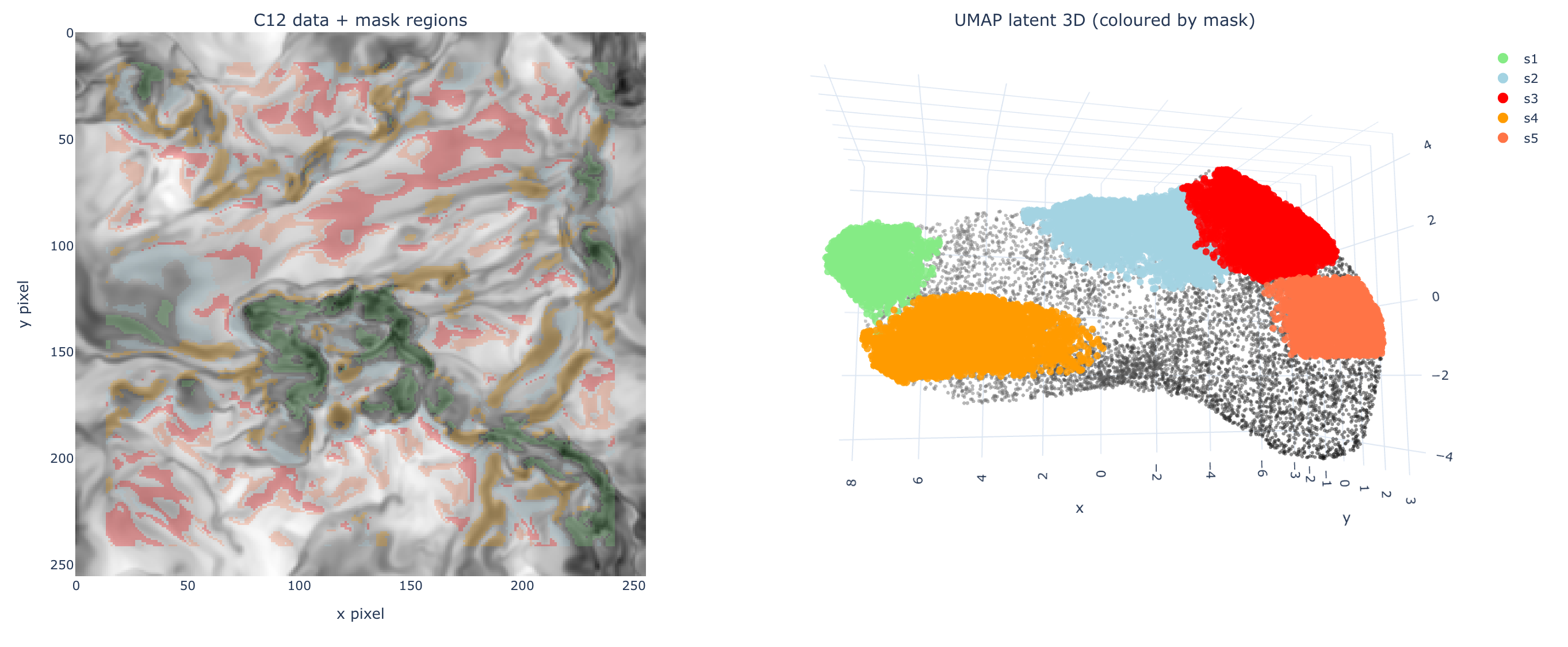}
\caption{ \textbf{Back-mapping representative latent neighborhoods in the MHD density field.} Left: the input density map with selected latent groups overlaid at their original spatial locations. Right: the corresponding three-dimensional UMAP projection of target-encoder embeddings, with the same groups highlighted in matching colors (rendered with the default perspective projection rather than orthographic, so the 3D point-cloud structure is easier to read). The selected neighborhoods occupy localized regions of latent space and map back to coherent density morphologies, including diffuse void-like areas, filamentary and interface-like structures, and compact dense clumps. These groups are representative latent selections rather than supervised classes or exhaustive segmentations.}
\label{fig:mhd_backmapping}
\end{figure}

To determine what this organization represents, we back-map selected UMAP
neighborhoods to the original density field
(Figure~\ref{fig:mhd_backmapping}). Localized regions of latent space correspond
to coherent spatial morphology, including filamentary interfaces, ridge-like
structures, compact high-contrast regions and diffuse material. These
neighborhoods are neither supervised classes nor exhaustive segmentations.
Instead, they are local regions of a continuous coordinate system whose
structure can be inspected directly in the original field.

The physical interpretation is validated by comparison with the known spatial
segregation of Alfv\'enic regimes in compressive MHD turbulence.
\citet{LiZhao2025mhd} showed that the local Alfv\'en Mach number,
${\cal M}_{\rm A}=\sqrt{E_k/E_B}$, distinguishes magnetically regulated,
magnetic--kinetic transition and kinetically dominated regimes. The magnetic
regime preferentially occupies lower-density gas, whereas the kinetic regime
preferentially occupies higher-density gas. In the present back-maps, these trends are
compared with void-like and filamentary density morphology, respectively.

ScaleAware-JEPA receives only the scalar density field, yet its latent
neighborhoods recover morphological distinctions aligned with this independently
established organization. The result indicates that density alone retains
structural information connected to magnetic regulation, and that the learned
latent atlas makes this information inspectable without prescribing diagnostic
thresholds or segmentation rules. The detailed physical interpretation of these
neighborhoods is developed in a dedicated domain-specific study.

\section*{Latent Atlases Across Distinct Field Regimes}

\begin{figure}[H]
\centering
\includegraphics[width=\textwidth]{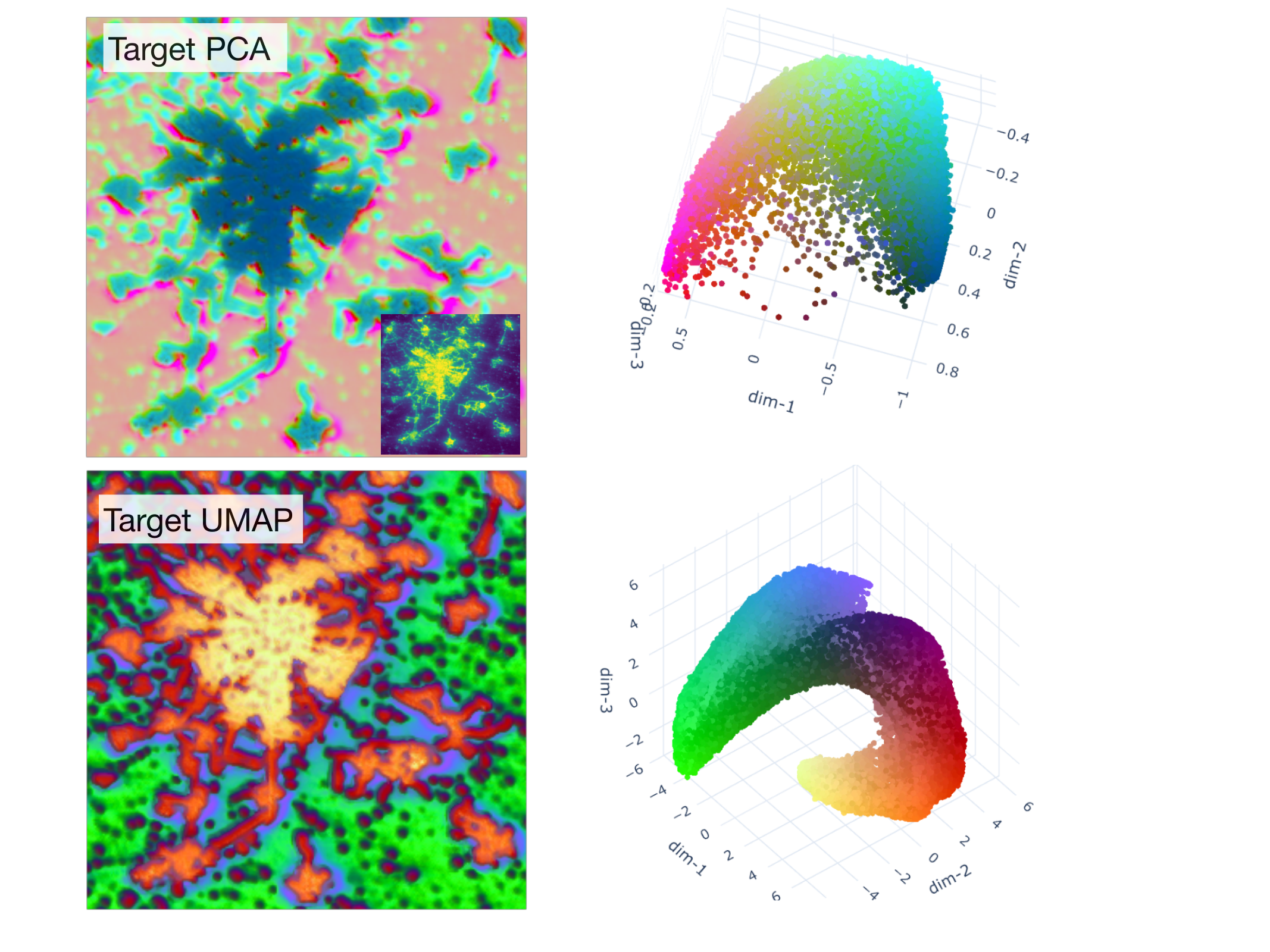}
\caption{\textbf{Dense latent topology learned for the Chengdu nighttime-light field.} Target-encoder embeddings are projected with PCA and UMAP and mapped back to their original spatial locations. The projections reveal coherent latent organization across the urban field, separating compact high-intensity cores, extended road-like structures, diffuse emission, and low-intensity surrounding regions. The inset shows the original nighttime-light image, providing the spatial reference for the latent back-mapping.}
\label{fig:jepa_chengdu}
\end{figure}

\begin{figure}[H]
\centering
\includegraphics[width=\textwidth]{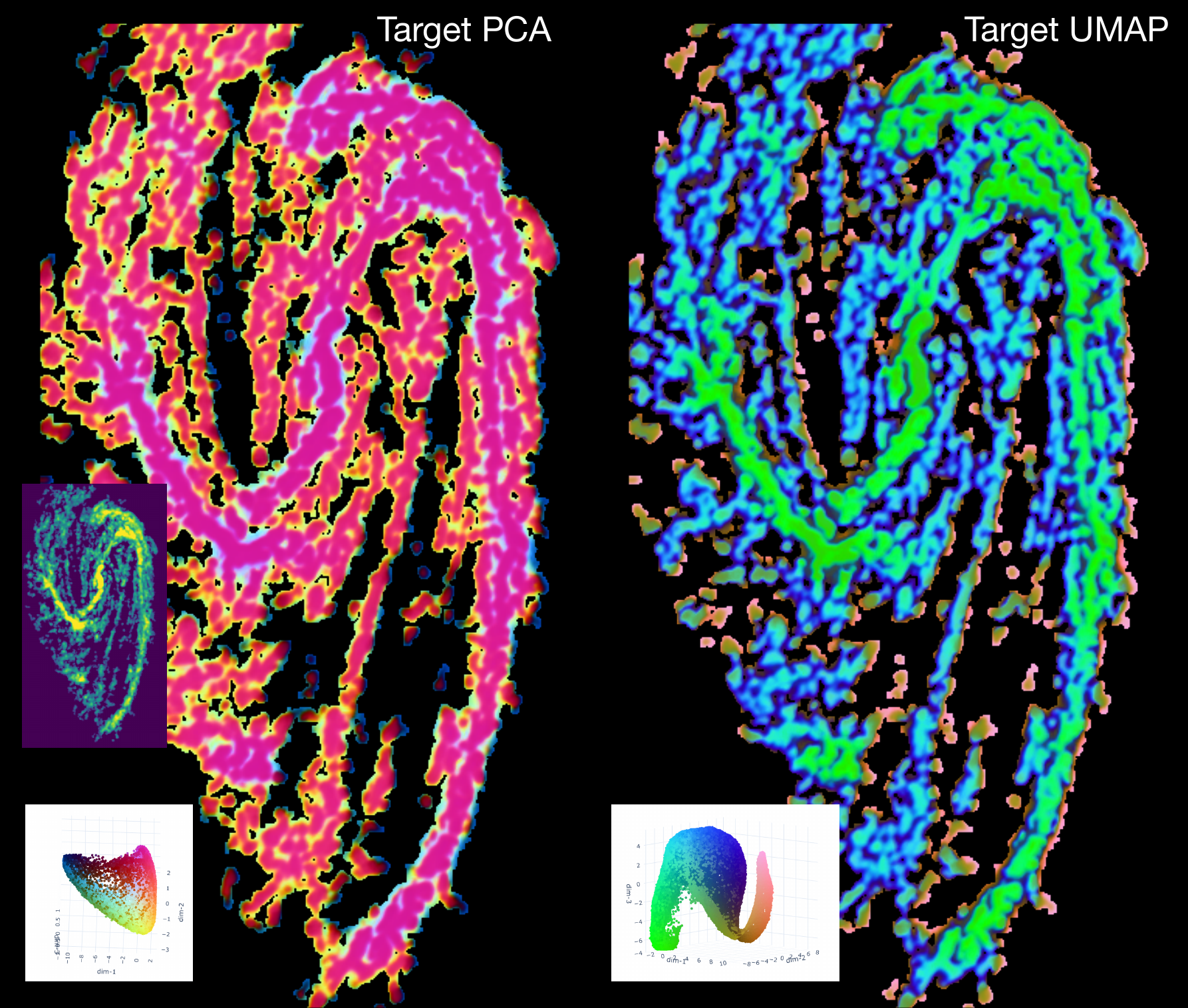}
\caption{\textbf{Dense latent topology learned for the NGC~3627 molecular-gas field.}
Target-encoder embeddings of the PHANGS--ALMA CO field are projected with PCA
and UMAP and then mapped back to their original spatial locations. The PCA
projection captures large-scale latent variation across the molecular disk,
separating the bright central concentration and spiral-arm molecular gas from
lower-surface-brightness interarm emission. The UMAP projection further
separates nonlinear structural neighborhoods: bright arm clouds, interarm
molecular clouds, diffuse extended emission, and narrow elongated interarm or
contrail-like molecular features occupy different regions of latent space. Some of the
narrow molecular structures are particularly relevant in light of the
galactic-scale molecular contrail reported in NGC~3627 by
\citet{ZhaoLi2025contrail}. The inset point clouds show that the representation
forms a continuous but structured manifold rather than a set of discrete
supervised classes. The separation is obtained without cloud labels, arm/interarm
masks, or contrail annotations, indicating that the self-supervised latent
coordinates organize physically meaningful molecular-gas morphology.}
\label{fig:jepa_ngc}
\end{figure}

Figures~\ref{fig:jepa_chengdu} and~\ref{fig:jepa_ngc} examine ScaleAware-JEPA
in two further scalar-field regimes: nighttime-light structure in Chengdu and
molecular-gas emission in the nearby galaxy NGC~3627. These fields differ from
simulated MHD turbulence in origin, dynamics and interpretation, but share the
same representational problem: meaningful morphology is distributed across
space and scale, without a complete set of predefined labels. The purpose is
therefore not to provide exhaustive domain-specific analyses, but to determine
whether the learned dense latent coordinates remain spatially interpretable
across distinct field-generating systems.

\paragraph{Nighttime lights in Chengdu.}
We examine a nighttime-light map of Chengdu derived from the NASA Black Marble
VIIRS product \citep{Roman2018BlackMarble}
(Figure~\ref{fig:jepa_chengdu}). Unlike the turbulent and astronomical fields
considered elsewhere, this field records the spatial organization of urban
activity rather than fluid or gravitational dynamics. Its inclusion tests
whether the latent-atlas construction depends on a particular physical
mechanism. The learned embedding organizes compact bright urban regions,
elongated light corridors, patchy suburban structure and low-intensity
background into distinct local neighborhoods, each of which can be mapped back
to coherent morphology in the original field.

\paragraph{Molecular gas in NGC 3627.}

For NGC~3627, we use the
PHANGS--ALMA CO(2--1) molecular-gas map
\citep{Leroy2021PHANGSALMA,Leroy2021PHANGSPipeline}. The learned representation
organizes the CO field along its dominant environmental structure without
labels, arm masks, or intensity thresholds. The PCA back-map captures the broad
contrast between the concentrated central and spiral-arm molecular disk and the
fainter diffuse interarm component, while the UMAP back-map resolves finer local
differences within this continuum. The embedding is thus sensitive to the
distinction between concentrated and diffuse molecular morphology. This is relevant to the recently reported molecular contrail in NGC~3627
\citep{ZhaoLi2025contrail}, a narrow extended CO structure distinct from the
ordinary bright arm population in the present latent map. Full
isolation of the kiloparsec-scale contrail as a single global object is limited
by the local ConvNeXt block footprint; what is recovered is the local
morphological character that sets it apart from its surroundings
(see Appendix~\ref{app:run_sweeps} for architecture details).

\section*{Conclusion}

Scientific fields are intrinsically multiscale, and a representation intended to capture their organization should be governed by that physical hierarchy rather than by the arbitrary geometry of a fixed image grid. ScaleAware-JEPA implements this principle through two coupled choices. CDD supplies a dense encoder with localized, pixel-registered components that expose fine, intermediate, and coarse morphology as aligned input structure, rather than requiring the full scale hierarchy to emerge indirectly from local convolutional processing. The same hierarchy determines the masking footprint, so latent prediction is posed with context matched to the physical scale of the hidden structure. CDD therefore provides the common geometry for both representation and prediction, while reducing the influence of oscillatory responses from conventional multiscale bases that can offer shortcuts to a predictive network.

Because its input representation and predictive task are defined by the field's
scale hierarchy rather than by domain-specific object definitions, label sets,
catalogues, or segmentation rules, ScaleAware-JEPA is applicable across systems
whose organization is nested across scale. MHD turbulence, urban nighttime-light
structure, and molecular-gas emission in NGC~3627 serve as deliberately
disparate demonstrations. In each setting, the framework produces a dense,
spatially back-mappable latent atlas in which diffuse regions, filamentary
interfaces, compact clumps, and faint extended structures occupy coherent
neighborhoods.

These results point to a design principle that is crucial for self-supervised
learning on multiscale systems: the representation and the predictive task
should be organized by the same physical scale hierarchy. The MHD sweep further shows that scale-aware masking provides a controlled
relationship between context footprint and latent-use diagnostics, whereas
fixed-box and excessively large masks approach hinge saturation. Scale-aware encoding makes multiscale structure available to
the network; scale-informed masking asks the network to predict that
structure at the scale on which it is organized.
The method also provides a direct route for domain scientists to incorporate
physical knowledge: the number of CDD scales and the mask-footprint multiplier
are interpretable parameters that express assumptions about the scale
structure of the field under study. This moves JEPA beyond fixed-patch visual
learning and toward a general architecture for latent extraction in complex
multiscale systems.

Code is available at \url{https://github.com/gxli/SA-JEPA}.

\section*{Acknowledgments}
The author received no external funding for this work.

\section*{Data Availability}
The datasets used in this work are publicly available from the sources cited in
the manuscript.

\section*{Resource Acknowledgment}
All experiments in this work were performed on self-funded consumer-grade
hardware, primarily an Apple MacBook Pro with an M3 Pro chip and a headless desktop
equipped with an NVIDIA RTX 3090 GPU.

\section*{Author Contributions}
GXL designed the project, wrote the code, performed the experiments and wrote the manuscript.
\bibliography{references}

\appendix

\section{Detailed CDD coordinates}

Differences between exponentially spaced diffusion times yield a
pixel-registered scale pyramid
\[
\mathcal{P}(x)=\{x^{(1)},\dots,x^{(S)}\}.
\]
The reported experiments use CDD scale coordinates $[2,4,8,16,32]$ (MHD,
Chengdu, and NGC). These scales are used both for CDD
extraction and for construction of the scale-dependent masks.

The CDD components are supplied to the encoder as a pixel-registered
multiscale stack together with a per-channel scale code that preserves scale
identity before cross-scale fusion. The same diffusion scales define the masking
geometry, so the context footprint for each component is tied to its physical
scale rather than to a single fixed image-space resolution.

Wavelet decompositions can introduce ringing and sign-changing lobes near sharp
structures, producing local, high-contrast and scale-correlated artifacts that
a predictor may exploit as shortcuts
\citep{li2022constrained,han2019gibbs}. CDD's
diffusion-based primitives are localized and pixel-registered, reducing this
confound and keeping latent prediction tied to field morphology.

\subsection*{Wavelet frontend control}
\label{app:cdd_wavelet_control}

As a frontend control, we trained matched MHD runs using a log-normal
wavelet decomposition in place of the constrained CDD
pyramid \citep{daubechies1992ten,mallat2009wavelet}. These runs remained
non-collapsed and produced moderate-to-high effective rank, but their hinge
dynamics were less well conditioned than the constrained baseline. Across mask
scales from $0.8\times\sigma_s$ to $1.6\times\sigma_s$, the hinge ratio rose
from $0.330$ to $0.993$. Thus the wavelet frontend yields a non-collapsed,
moderate-rank representation, but the spread hinge saturates at larger mask
scales, whereas constrained CDD maintains headroom across the full sweep. The
$1.2\times\sigma_s$ run is selected as the visualization representative
(Table~\ref{tab:wavelet_frontend_diagnostic}).

\begin{figure}[H]
\centering
\includegraphics[width=\textwidth]{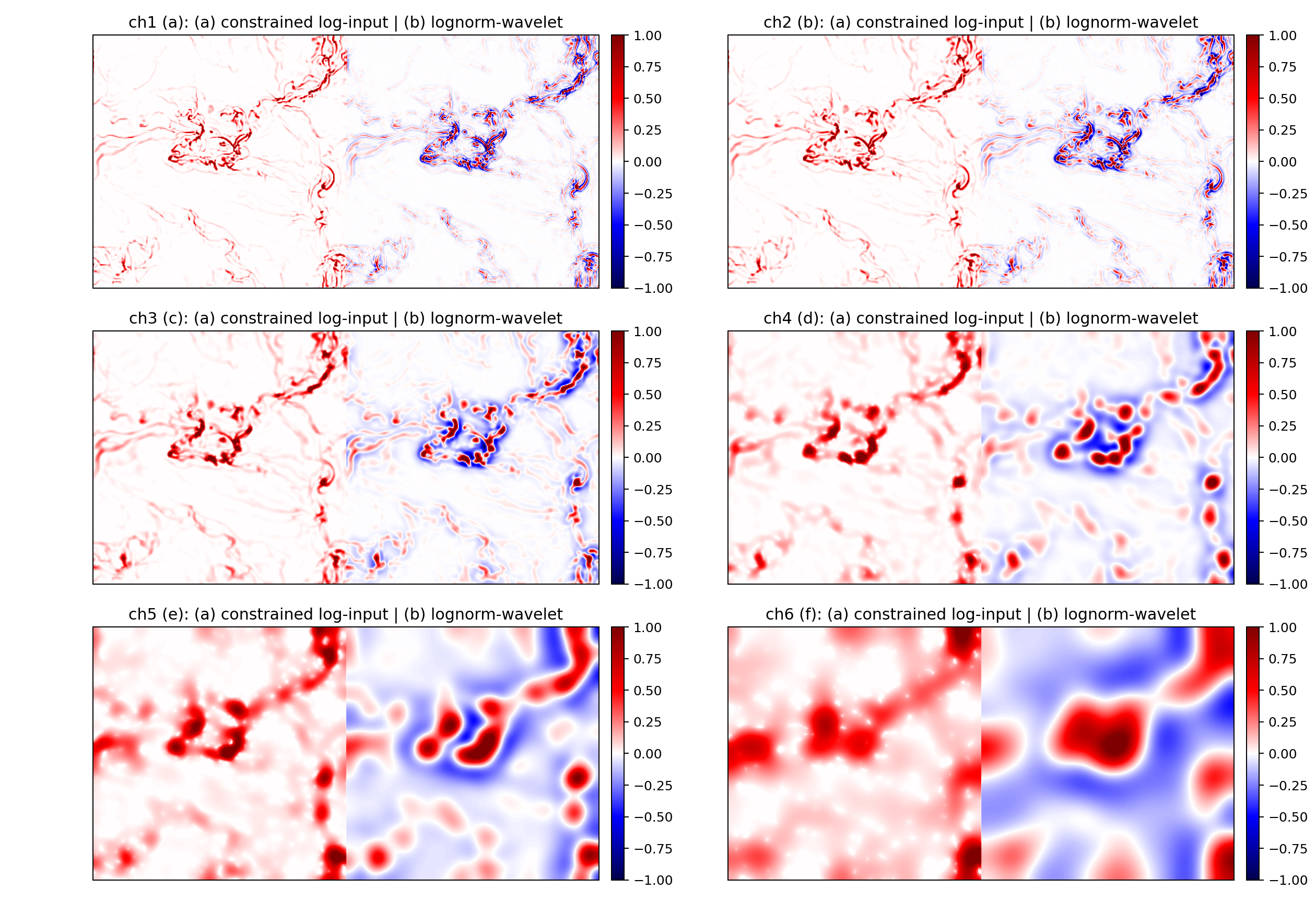}
\caption{\textbf{CDD versus log-normal wavelet decomposition.}
Each panel compares the CDD primitive (left half) with the matched log-normal
wavelet coefficient (right half) across six scales on an MHD turbulence field.
CDD produces sparse, spatially compact features with near-zero response in
smooth regions; the wavelet representation exhibits diffuse oscillatory leakage
and ringing artefacts that are absent in the CDD primitives.}
\label{fig:cdd_vs_wavelet}
\end{figure}

Figure~\ref{fig:cdd_vs_wavelet} compares the CDD and wavelet decompositions
directly across six scales on an MHD field. At fine scales, CDD produces
near-zero response in smooth inter-filament regions, whereas wavelet
coefficients exhibit diffuse oscillatory leakage. At intermediate scales, CDD
isolates filamentary structures as sparse, spatially compact features; the
wavelet representation smears the same structures into broad, overlapping halos.
At coarse scales, both methods recover large-scale topology, but wavelet
coefficients retain visible ringing artefacts absent in the CDD primitives
(see Figure~\ref{fig:cdd_vs_wavelet} and Table~\ref{tab:wavelet_frontend_diagnostic}).

Figure~\ref{fig:wavelet_frontend_control} shows the corresponding wavelet
latent map. The UMAP manifold is organized, confirming that the run is not
collapsed, but the mapped-back RGB field exhibits broad halo-like bands around
high-contrast structures---a shadow pattern reminiscent of the negative
oscillatory signal visible in the wavelet side of
Figure~\ref{fig:cdd_vs_wavelet}. This behavior, consistent with oscillatory cross-scale
leakage and halo-like responses near sharp transitions \citep{han2019gibbs},
indicates that the wavelet frontend is less localized but not collapsed. 


\begin{center}
\includegraphics[width=\textwidth]{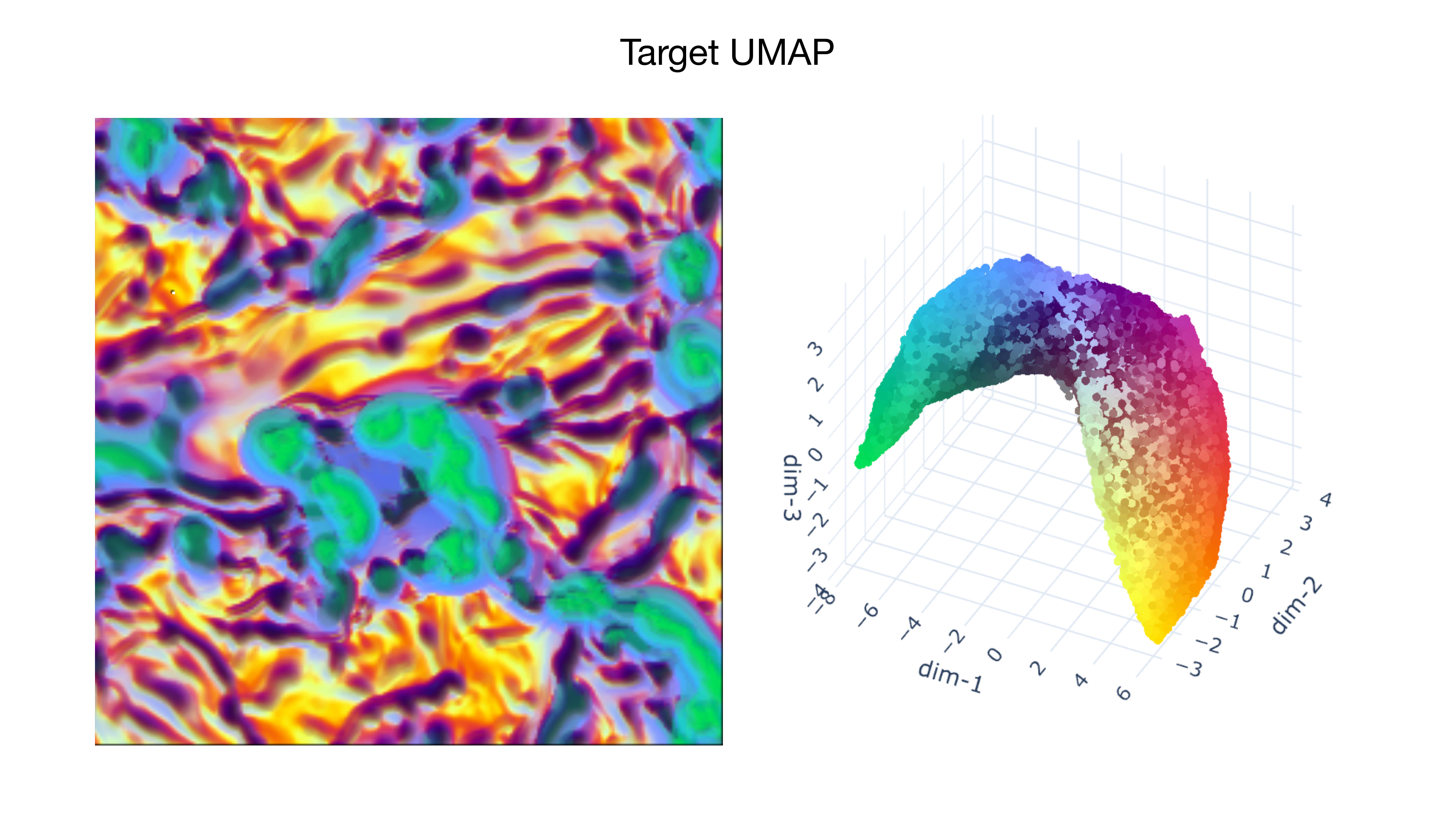}
\captionof{figure}{\textbf{Wavelet-frontend control on the MHD field.}
A matched JEPA run using a wavelet frontend produces an organized
UMAP manifold, but the mapped-back UMAP RGB field shows broad halo-like bands
around high-contrast structures. This indicates a non-collapsed but less
localized frontend behavior, consistent with oscillatory leakage in the wavelet
decomposition and with ringing near sharp transitions.}
\label{fig:wavelet_frontend_control}
\end{center}

\begin{table}[H]
\centering
\caption{\textbf{Wavelet-frontend MHD diagnostics.}
Matched MHD runs using a wavelet frontend
remain non-collapsed and produce moderate-to-high effective rank, but the
hinge ratio rises toward saturation at larger pyramid mask scales, whereas
constrained CDD maintains headroom across the sweep. The $1.2\times\sigma_s$ run is selected for visualization.}
\label{tab:wavelet_frontend_diagnostic}
\begin{tabular}{rrrr}
\toprule
Mask scale & Target rank & Pred. rank & Hinge ratio \\
\midrule
$0.8\times\sigma_s$        & 2.199 & 1.902 & 0.330 \\
$1.2\times\sigma_s^\ast$ & \textbf{2.938} & \textbf{4.316} & 0.609 \\
$1.6\times\sigma_s$        & 3.334 & 2.824 & 0.993 \\
\bottomrule
\end{tabular}
\end{table}

\section{Scale-Aware Encoder Implementation}
\label{app:scale_aware_encoder}

The CDD decomposition is treated as a separate preprocessing step that produces
a pixel-registered pyramid of scale components
$\mathcal{P}(x)=\{x^{(1)},\ldots,x^{(S)}\}$ (see Figure~\ref{fig:framework}
for the architectural overview). Unlike a conventional feature pyramid, these
levels are not produced by progressive downsampling inside the network. They
are obtained before learning, and all remain registered at the input
resolution.

For each scale $s$, the encoder forms a per-scale input by stacking the CDD
component $x^{(s)}$, the corresponding mask channel $m^{(s)}$, and a scalar
scale code $c^{(s)}$. A shared scale-adapter architecture maps this input into a
common feature width,
\[
h^{(s)} = A_\theta\!\left(x^{(s)}, m^{(s)}, c^{(s)}\right),
\qquad s=1,\ldots,S .
\]
The adapter therefore sees not only the CDD value but also which pixels were
hidden and which diffusion scale the channel represents. This gives each scale
local spatial processing before cross-scale fusion and places all CDD components
into a comparable feature space.

The adapted features are fused across scale,
\[
h = \operatorname{Fuse}\left(h^{(1)},\ldots,h^{(S)}\right).
\]
After the shared per-scale adapter, scale features are fused by a top-down
residual pathway: coarse features are successively added to finer-scale
features, and each fused scale is passed through its own $1\times1$
projection before concatenation and dense ConvNeXt processing. This differs
from a standard feature pyramid because the hierarchy is supplied by CDD
before learning rather than generated by neural downsampling.

The fused multiscale feature map is processed by a dense ConvNeXt backbone
\citep{liu2022convnext,woo2023convnextv2},
\[
z = F_\theta(h),
\]
which performs spatial mixing without reducing the field resolution. The same
encoder architecture serves the online context branch and the EMA target branch.
Default implementation details are listed in
Table~\ref{tab:scale_aware_encoder_defaults}.

Several normalization layers stabilize training. Per-scale adapter normalization
prevents high-amplitude CDD channels from dominating the shared adapter. Stem
normalization stabilizes the first ConvNeXt projection after scale fusion. A
final LayerNorm after the encoder head places the dense latent vectors on a
common scale before prediction and analysis.

\begin{table}[H]
\centering
\caption{\textbf{Default scale-aware ConvNeXt encoder settings.}
Implementation defaults used in the reported experiments, grouped by functional role.}
\label{tab:scale_aware_encoder_defaults}
\begin{tabular}{lp{0.52\columnwidth}}
\toprule
Component & Default setting \\
\midrule
\multicolumn{2}{l}{\textit{Input and scale representation}} \\
Input representation & CDD pyramid \\
CDD scales & $[2,4,8,16,32]$ \\
Per-scale adapter input & field, mask channel, scale code \\
\midrule
\multicolumn{2}{l}{\textit{Per-scale adaptation}} \\
Adapter sharing & shared architecture across scales \\
Adapter kernel & $3\times3$ \\
Adapter feature width & $8$ channels per scale \\
\midrule
\multicolumn{2}{l}{\textit{Cross-scale fusion}} \\
Scale fusion & top-down residual fusion; per-scale $1\times1$ projection \\
Spatial downsampling & none \\
\midrule
\multicolumn{2}{l}{\textit{Dense ConvNeXt backbone}} \\
Encoder backbone & dense ConvNeXt \\
Encoder width & $64$ channels \\
Encoder depth & $4$ blocks \\
ConvNeXt spatial kernel & $7\times7$ depthwise convolution \\
Global response normalization & enabled \\
\midrule
\multicolumn{2}{l}{\textit{Latent output}} \\
Latent channels & $32$ \\
Final normalization & LayerNorm \\
Output & dense latent map $z\in\mathbb{R}^{32\times H\times W}$ \\
\bottomrule
\end{tabular}
\end{table}

\section{Mask construction}
\label{app:mask_construction}

For channel $s$, the mask footprint is
\begin{equation}
b = \max(n_{\rm target},\; \operatorname{round}(\sigma_s f_{\mathrm{mask}} + B_0)),
\;
B_s =
\begin{cases}
\operatorname{oddceil}(b), & b \le B_{\mathrm{cap}},\\[4pt]
\operatorname{oddfloor}(B_{\mathrm{cap}}), & b > B_{\mathrm{cap}},
\end{cases}
\end{equation}
where $n_{\rm target}=3$, $\operatorname{oddceil}(x)$ is the smallest odd
integer not smaller than $x$, $\operatorname{oddfloor}(x)$ the largest odd
integer not larger than $x$, $f_{\mathrm{mask}}$ is the scale multiplier,
$B_0$ is a fixed offset, and $B_{\mathrm{cap}}=48$\,px (MHD, Chengdu) or
$35$\,px (NGC). Setting $f_{\mathrm{mask}}=0$ recovers a fixed-box mask;
$B_0=0$ gives a pure scale-tied pyramid mask. When the hard cap is active
the box is odd-rounded downward to avoid exceeding $B_{\mathrm{cap}}$.
The mask remains centered on the target and always covers the prediction
patch. The final CDD channel contains the coarsest scale together with the
unresolved residual.

For each sampled target center, the corresponding $B_s\times B_s$ region is
removed from the context component $x_c^{(s)}$. A separate binary
mask-indicator channel marks the removed pixels. The prediction loss is
evaluated only on the central $3\times3$ target patch. Increasing $B_s$ therefore changes the spatial context available
for prediction without changing the size of the localized target used in the
loss.

For the MHD sweep, we use effective rank together with the hinge ratio
\begin{equation}
r_{\mathrm{hinge}}
=
\frac{\overline{\mathcal{L}}_{\mathrm{spread}}^{\mathrm{late}}}
{\overline{\mathcal{L}}_{\mathrm{spread}}^{\mathrm{early}}+\epsilon},
\end{equation}
where the numerator and denominator are averages over late- and early-training
windows. Values near zero indicate substantial hinge decay, whereas values near
unity indicate a plateau. An excessively large mask can cause context
starvation: the model receives insufficient information to reduce the
batch-wise variance constraint, and increasing the footprint no longer improves
latent use in a controlled way.

\section{Training objective}
\label{app:training_config}

Let $f_\theta$ be the online encoder, $f_{\bar{\theta}}$ the EMA target
encoder, $p_\theta$ and $p_{\bar{\theta}}$ the corresponding pointwise
projectors, and $g_\theta$ a spatial predictor. For a masked context view
$x_c$ and unmasked target view $x_t$,
\begin{equation}
q_c = p_\theta(f_\theta(x_c)), \qquad
q_t = p_{\bar{\theta}}(f_{\bar{\theta}}(x_t)), \qquad
\hat q_t = g_\theta(q_c).
\end{equation}
The target encoder is updated by exponential moving average,
\begin{equation}
\bar{\theta} \leftarrow m\bar{\theta} + (1-m)\theta.
\end{equation}

For valid target patches indexed by $(b,k)\in\Omega$, projected patch tensors
$\hat Q_{bk},Q_{bk}\in\mathbb{R}^{C\times P\times P}$ are spatially pooled:
\[
\hat q_{bk} = \frac{1}{P^2}\sum_{i,j=1}^{P}\hat Q_{bk,:,i,j},
\qquad
q_{bk} = \frac{1}{P^2}\sum_{i,j=1}^{P} Q_{bk,:,i,j}.
\]
The prediction loss is
\begin{equation}
\mathcal{L}_{\mathrm{pred}}
=
\frac{1}{|\Omega|\,C}
\sum_{(b,k)\in\Omega}
\bigl\|\hat q_{bk}-q_{bk}\bigr\|_2^2 .
\end{equation}
For 3D patches, pooling is performed over $P^3$ voxels.

To prevent collapse, we apply a weak spread regularizer to projected context
tokens $u_i\in\mathbb{R}^{C}$. After mean centering,
\begin{equation}
\tilde{u}_i = u_i - \frac{1}{N}\sum_{j=1}^{N}u_j ,
\end{equation}
the population standard deviation in channel $c$ is
\begin{equation}
\sigma_c =
\left(
\frac{1}{N}\sum_{i=1}^{N}\tilde{u}_{i,c}^{\,2}+\epsilon
\right)^{1/2},
\end{equation}
and the spread term is
\begin{equation}
\mathcal{L}_{\mathrm{spread}}
=
\frac{1}{C}\sum_{c=1}^{C}\max(0,\tau-\sigma_c).
\end{equation}

The full objective is
\begin{equation}
\mathcal{L}
=
\lambda_{\mathrm{pred}}\mathcal{L}_{\mathrm{pred}}
+
\lambda_{\mathrm{spread}}\mathcal{L}_{\mathrm{spread}} .
\end{equation}

The selected configuration uses
$\lambda_{\mathrm{pred}}=50$,
$\lambda_{\mathrm{spread}}=5$, and
$\tau=1$.
Detailed projector, predictor, optimization and inference settings are tabulated
in the appendix.

\section{Training Configuration (repository defaults)}
\label{app:training_defaults}

This section records the configurable repository defaults. All settings
listed below can be overridden through the configuration system. The
manuscript reports separate selected-run settings where applicable; sweep
results and run selection are given in Appendix~\ref{app:run_sweeps}.

\begin{table}[H]
\centering
\caption{\textbf{Optimization, inference, and diagnostic defaults (repository).}
All settings are configurable; selected-run overrides are noted in the sweeps appendix.}
\label{tab:training_runtime_defaults}
\begin{tabular}{lp{0.55\columnwidth}}
\toprule
Setting & Default \\
\midrule
\texttt{normalize\_loss\_l2} & disabled \\
\texttt{scaleaware\_final\_norm} & LayerNorm, enabled \\
Projector & online and EMA; $1{\times}1(32\!\rightarrow\!96) \rightarrow \text{LayerNorm} \rightarrow
\text{GELU} \rightarrow 1{\times}1(96\!\rightarrow\!32)$ \\
Regularizer input & valid context patches after online projector (pooled spatial mode) \\
Prediction target & EMA target branch after target projector \\
Predictor & $1{\times}1 \rightarrow \text{LayerNorm} \rightarrow \text{GELU} \rightarrow 3{\times}3
\text{(reflect)} \rightarrow \text{LayerNorm} \rightarrow \text{GELU} \rightarrow 1{\times}1$ \\
\texttt{predictor\_hidden} & $96$ \\
Encoder depth & $4$ \\
Dilations & $[1,1,1,1]$ (MHD, Chengdu); $[1,1,2,4]$ (NGC) \\
\midrule
Epochs & $10$ \\
Batch size & $4$ (MHD); $1$ (NGC, Chengdu) \\
Optimizer & AdamW, lr $10^{-4}$, min $10^{-6}$, warmup $1.0$\,epoch \\
Weight decay & $10^{-5}$ \\
EMA momentum & $0.99 \rightarrow 0.9999$, warmup fraction $0.25$ \\
\midrule
Prediction weight & $50$ \\
Spread weight & $5$ ($\tau\!=\!1$, context, pooled) \\
\midrule
Inference TTA & available (\texttt{flip4}); not used for selected runs \\
Diagnostics & effective-rank, scale-response probe \\
\bottomrule
\end{tabular}
\end{table}

\subsection*{Data and Preprocessing}
\label{app:data_preprocessing}

The input file pattern is dataset-specific. For the MHD experiments, we use a
two-dimensional density slice from the selected simulation run.
For MHD training, we apply D4 augmentations (rotations and flips that
preserve statistical isotropy); for Chengdu and NGC, only flip
augmentations are used. During inference, masking is disabled
and the frozen target encoder is applied to the full field. Flip-based
test-time augmentation (not used for the selected runs) is available for the final dense latent maps.

\subsection*{Mask Parameterization and Target Sampling}
\label{app:mask_target_sampling}

The mask footprint is defined by the unified equation in
Section~\ref{app:mask_construction} (see also the main text,
Section~\ref{sec:multiscale_masking}). Target centers are generated by
random sampling with overlap rejection: a proposed target is rejected if
it overlaps an already accepted
target under the configured non-overlap rule. This produces spatially dispersed
target regions while preserving stochastic target placement across batches. In
the selected vanilla runs, overlap rejection is enabled through
\texttt{target\_nonoverlap=true}. Additionally, proposed targets whose centers
lie within half the encoder's local convolutional footprint of the field
boundary are rejected, so that masked context regions never extend beyond the
field edge and predictions are not contaminated by padding or missing data.

The default configuration uses random target sampling with overlap rejection.

\begin{table}[H]
\centering
\caption{\textbf{Masking and target-sampling defaults (repository).}
All settings are configurable. Sweep results are reported separately.}
\label{tab:training_masking_sampling}
\begin{tabular}{lp{0.55\columnwidth}}
\toprule
Setting & Default / role \\
\midrule
\texttt{mask\_size\_scaling} & scale-tied footprint multiplier $f_{\mathrm{mask}}$; default $1.0$ \\
\texttt{mask\_size} & fixed footprint $B_0$ in pixels; default $0$ (scale-based); overridden in box sweeps \\
\texttt{mask\_box\_hardcap} & configurable; manuscript selected runs use $48$\,px (MHD, Chengdu), $35$\,px (NGC) \\
\texttt{target\_sampling\_mode} & \texttt{random}, with overlap rejection \\
\texttt{target\_nonoverlap} & \texttt{true}; proposed targets overlapping accepted targets are rejected \\
\bottomrule
\end{tabular}
\end{table}

\subsection*{Loss Terms}
\label{app:loss_terms}

The final training objective is dominated by the JEPA prediction loss. Predicted
and target embeddings are compared only at masked target locations. For the selected vanilla runs, patch embeddings are compared without L2 normalizing the prediction loss, so latent amplitude remains available to the predictive objective.

A weak context spread regularizer prevents collapse by penalizing latent
channels whose batch standard deviation falls below the target value
$\tau=1$. The loss terms are summarized in Table~\ref{tab:training_runtime_defaults}.

\subsection*{Optimization, Inference, and Diagnostics}
\label{app:optimization_inference_diagnostics}

All final runs use AdamW \citep{loshchilov2019decoupled} with cosine
learning-rate decay and EMA target-encoder updates. {The online encoder and
context projector are optimized by backpropagation, while the target encoder and
target projector are updated only by exponential moving average. The encoder
produces a dense 32-channel latent map. In the selected configuration, the
projector is enabled and maps this representation to a 96-channel projected
space before the predictor and before the spread regularizer. The predictor is a
compact full-resolution convolutional head with hidden width 96, predictor
LayerNorm enabled, and a reflect-padded spatial $3{\times}3$ convolution.}

During inference, masking is disabled and the frozen target branch is applied
densely to the full field. Flip-based test-time augmentation (not used for the selected runs) is
available for the final latent maps. We compute effective-rank diagnostics as collapse and
latent-use checks, and we enable the scale-response probe to measure how
strongly the trained encoder depends on each CDD input channel. These
diagnostics are used to screen for non-collapse and latent-space usage; final
interpretation is based on the dense latent topology and mapped-back spatial
structures rather than on any scalar diagnostic alone.

\subsection*{Latent Map Visualization and Diagnostics}
\label{app:vis}

After training, masking is disabled and the frozen target encoder is applied to the full field to produce a dense latent map. For large fields, we evaluate overlapping windows and blend the outputs to reduce boundary artifacts. As during training, pixels within half the encoder's local convolutional footprint of the field boundary have reduced effective context; their latent coordinates may be less reliable than interior pixels and should be interpreted with caution.

For visualization, latent vectors are sampled from the dense map and projected with PCA or UMAP \citep{mcinnes2018umap}. PCA is used as a linear view of the global latent geometry, while UMAP is used as a nonlinear neighborhood-preserving view. {The UMAP
projection uses standardized input vectors with Euclidean metric,
$\texttt{min\_dist}=0.2$, $\texttt{n\_neighbors}=50$, and no L2 normalization
of the input embeddings ($\texttt{l2\_normalize}=\texttt{false}$).} The fitted
projection is then evaluated over the full spatial map and normalized for RGB
rendering.

We use effective rank and participation-style diagnostics only as collapse and latent-use checks. Low values indicate that the embedding has contracted to a small number of directions, while broader spectra indicate greater use of the latent space. These metrics are not treated as a standalone measure of physical quality; final interpretation comes from mapping latent neighborhoods and extremal regions back to the original field.

\subsection*{Symmetry consistency (optional)}

A weak symmetry consistency loss is available as an option to encourage
invariance under discrete field-preserving transformations. Four flip views are
used: identity, horizontal, vertical, and combined horizontal--vertical flip.
The context encoder is evaluated on each view of the same input; the resulting
feature maps are inverse-aligned, and the population variance across the
inverse-aligned views is averaged over batch, channels, and spatial positions:
\begin{equation}
    \mathcal{L}_{\mathrm{sym}}
    =
    \frac{1}{B\,G\,C\,H\,W}
    \sum_{b=1}^{B}
    \bigl\|\mathbf{F}_b - \bar{\mathbf{f}}_b\bigr\|_F^2,
\end{equation}
where $\mathbf{F}_b\in\mathbb{R}^{G\times C\times H\times W}$ stacks the
$G=4$ flip views, $\bar{\mathbf{f}}_b$ is the view-averaged tensor,
$\|\cdot\|_F$ is the Frobenius norm, $B$ is the batch size, $C$ the number
of latent channels, and $H,W$ the spatial dimensions of the encoder feature
map. The symmetry term is not used for the
selected runs reported in the main text but is available as an optional
regularizer.

\subsection*{Test-time augmentation (optional)}

During inference, masking is disabled and the frozen target branch is applied
densely to the full field. Flip-based test-time augmentation blends the same
four flip views used for symmetry training to produce the final dense latent
map. TTA is not used for the selected runs but is available to reduce 
asymmetry when needed.

\section{Masking Sweeps and Run Selection}
\label{app:run_sweeps}

Compact masking sweeps were used to select representative embeddings for the main visualizations, serving as stability and interpretability checks across masking regimes rather than searches for a universal optimal mask. Each candidate run was required to remain non-collapsed, use more than one latent direction, and produce a dense embedding whose neighborhoods map back to coherent spatial structures.

Collapse and latent-space usage were assessed through two diagnostics computed from the covariance spectrum with eigenvalues $\lambda_i$. Defining normalized eigenvalues $p_i = \lambda_i / \sum_j \lambda_j$, the effective rank is
\begin{equation}
r_{\mathrm{eff}} = \exp\!\left(-\sum_i p_i \log p_i\right),
\end{equation}
and the participation number is
\begin{equation}
r_{\mathrm{part}} = \frac{\left(\sum_i \lambda_i\right)^2}{\sum_i \lambda_i^2}.
\end{equation}
A dead channel is flagged when its standard deviation over sampled spatial embeddings falls below a fixed numerical threshold. These diagnostics identify collapse or excessive contraction but are not standalone measures of physical quality; final selection was based on visual inspection of PCA and UMAP maps and their mapped-back spatial structures, because higher effective rank did not always correspond to cleaner or more physically interpretable embeddings.

All selected runs use a CDD pyramid with the unresolved residual folded into the last scale channel: $[2,4,8,16,32]$ (MHD, Chengdu, NGC). The encoder uses four ConvNeXt blocks; dilations are $[1,1,1,1]$ for MHD and Chengdu, $[1,1,2,4]$ for NGC. The encoder's local convolutional receptive field is 25~px (49~px for NGC with dilated blocks); pixels within half this width of the field boundary lack full context and are therefore excluded from both target placement during training and latent evaluation during inference, with affected inference pixels set to NaN. A hard cap of $48$~px (MHD, Chengdu) or $35$~px (NGC) is applied to the masked context footprint. For pyramid masks, the footprint is expressed relative to the CDD diffusion scale $\sigma_s$ of each input channel; for box masks, it is a fixed image-space size in pixels. For NGC, more than 50\% of the field pixels are noise-dominated; target centers are restricted to regions with intensity exceeding $3.5\times$ the RMS noise, and inference is evaluated only within those valid regions.

\begin{figure}[H]
\centering

\begin{minipage}[t]{0.30\textwidth}
    \centering
    \includegraphics[width=\textwidth]{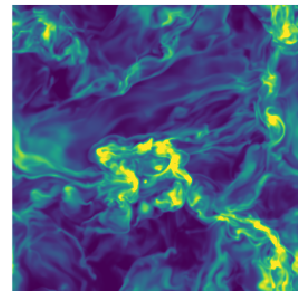}
\end{minipage}
\hfill
\begin{minipage}[t]{0.60\textwidth}
    \centering
    \includegraphics[width=\textwidth]{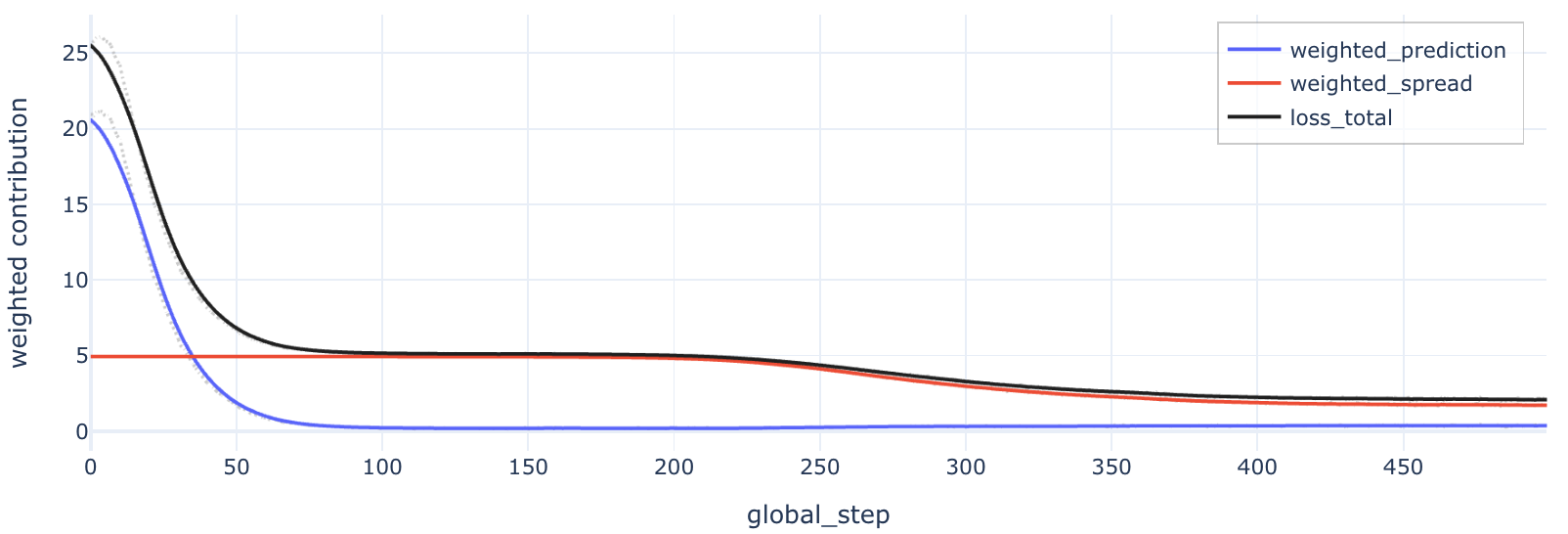}
\end{minipage}

\vspace{0.8em}

\begin{minipage}[t]{0.30\textwidth}
    \centering
    \includegraphics[width=\textwidth]{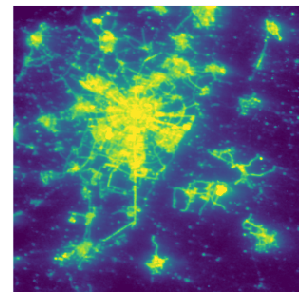}
\end{minipage}
\hfill
\begin{minipage}[t]{0.60\textwidth}
    \centering
    \includegraphics[width=\textwidth]{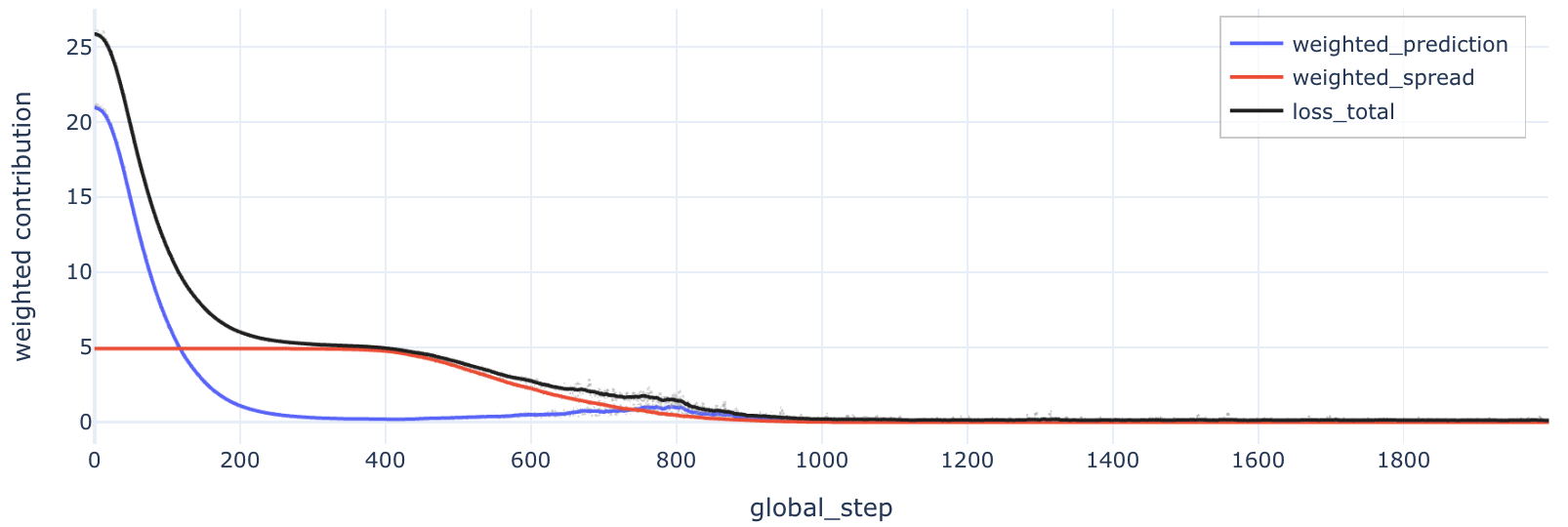}
\end{minipage}

\vspace{0.8em}

\begin{minipage}[t]{0.30\textwidth}
    \centering
    \includegraphics[width=\textwidth]{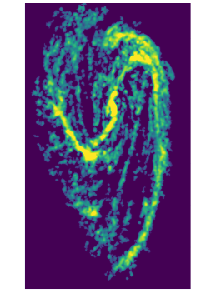}
\end{minipage}
\hfill
\begin{minipage}[t]{0.60\textwidth}
    \centering
    \includegraphics[width=\textwidth]{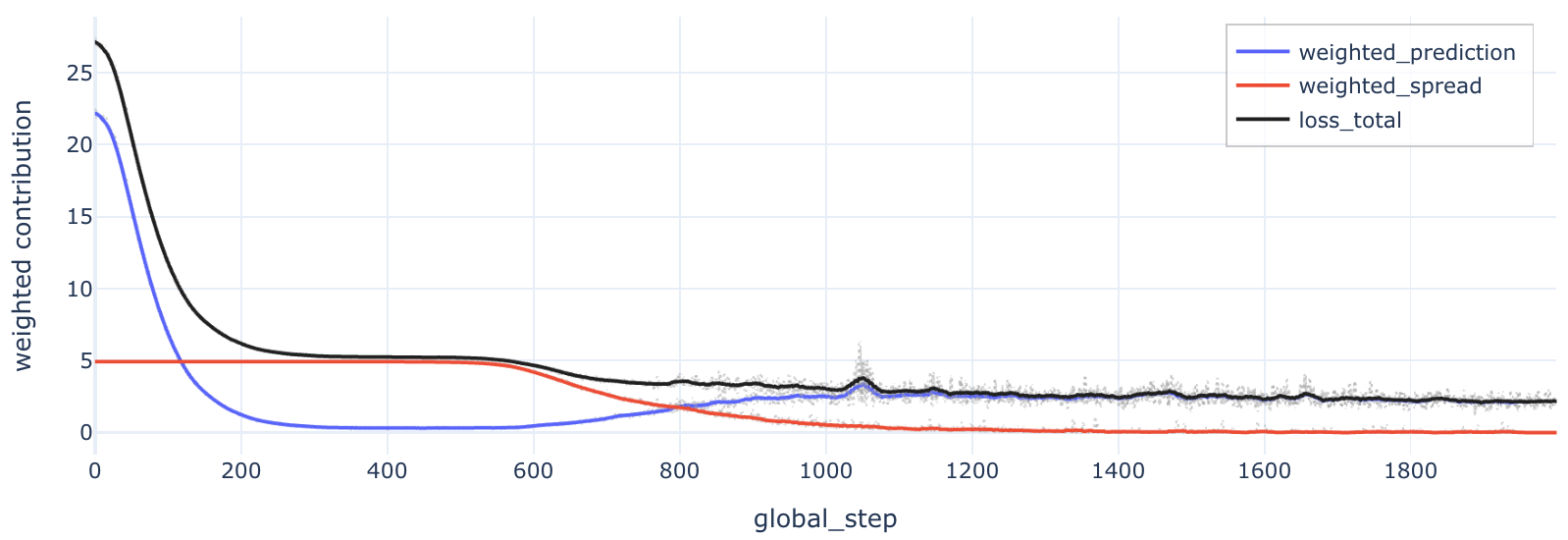}
\end{minipage}
\caption{\textbf{Selected-run input fields and training losses.}
For each selected vanilla run, the left panel shows the scalar input field and
the right panel shows the corresponding training-loss history. Rows correspond
to MHD turbulence, Chengdu nighttime lights, and the NGC molecular-gas field,
respectively. MHD uses pyramid masking with footprint $1.2\times\sigma_s$;
Chengdu uses $0.8\times\sigma_s$; NGC uses $1.6\times\sigma_s$, where $\sigma_s$
is the CDD diffusion scale of the corresponding input channel. These diagnostics show the optimization
behavior associated with the dense latent topologies shown in
Figures~\ref{fig:jepa_mhd}, \ref{fig:jepa_chengdu}, and~\ref{fig:jepa_ngc}.}
\label{fig:app_selected_data_loss}
\end{figure}

The selected runs are Chengdu pyramid scale~0.8, MHD pyramid scale~1.2, and NGC pyramid scale~1.6, marked in Table~\ref{tab:final_pooled_sweep_and_box_control}. In the MHD sweep, scale~1.2 lies in an intermediate transition regime of the latent-use diagnostics: larger pyramid and fixed-box masks can reach comparable or higher effective rank, but scale~1.2 gives the strongest balance among effective rank, predictor-side usage, and spatial interpretability. For Chengdu, scale~0.8 produces the highest target participation ratio (1.62) and clean structural separation under PCA and UMAP inspection. For NGC, scale~1.6 gives similarly clean separation of compact, extended, and diffuse molecular-gas structures. Training-loss histories for all three runs are shown in Figure~\ref{fig:app_selected_data_loss}; in each case the spread regularizer remains stable and the weighted prediction term provides the dominant training signal.

\begin{table}[H]
\centering
\caption{\textbf{Final pooled masking sweep and fixed-box MHD control.}
All runs use unnormalized latent prediction, patch-scale normalization, final
latent normalization, and a pooled standard-deviation hinge spread regularizer.
Reported values are final sampled diagnostics. Selected visualization runs are
marked with an asterisk.}
\label{tab:final_pooled_sweep_and_box_control}

\setlength{\tabcolsep}{5pt}
\renewcommand{\arraystretch}{1.05}

\begin{tabular}{l l r r r r r}
\toprule
Dataset & Strategy & Footprint & Target rank & Pred. rank & Target part. & Hinge ratio \\
\midrule

\textit{Chengdu} \\
\quad & pyramid & $0.8\times\sigma_s^\ast$ & \textbf{2.306} & \textbf{1.452} & \textbf{1.62} & 0.000 \\
\quad & pyramid & $1.2\times\sigma_s$      & 1.568 & 2.160 & 1.25 & 0.004 \\
\quad & pyramid & $1.6\times\sigma_s$      & 1.166 & 1.163 & 1.06 & 0.005 \\
\quad & pyramid & $2.0\times\sigma_s$      & 1.157 & 1.560 & 1.05 & 0.008 \\

\midrule
\textit{MHD} \\
\quad & pyramid & $0.4\times\sigma_s$      & 1.468 & 1.272 & 1.16 & 0.244 \\
\quad & pyramid & $0.8\times\sigma_s$      & 1.691 & 1.803 & 1.25 & 0.219 \\
\quad & pyramid & $1.2\times\sigma_s^\ast$ & \textbf{2.152} & \textbf{2.044} & \textbf{1.44} & 0.354 \\
\quad & pyramid & $1.6\times\sigma_s$      & 3.183 & 4.013 & 2.01 & 0.544 \\
\quad & pyramid & $2.0\times\sigma_s$      & 2.980 & 3.277 & 1.82 & 0.999 \\
\quad & box     & $7$ px                   & 3.099 & 3.842 & 1.86 & 0.946 \\
\quad & box     & $11$ px                  & 2.918 & 3.076 & 1.78 & 1.000 \\
\quad & box     & $15$ px                  & 2.913 & 3.065 & 1.78 & 1.000 \\
\quad & box     & $19$ px                  & 2.901 & 3.048 & 1.77 & 1.000 \\

\midrule
\textit{NGC} \\
\quad & pyramid & $0.8\times\sigma_s$      & 1.088 & 1.160 & 1.03 & 0.000 \\
\quad & pyramid & $1.2\times\sigma_s$      & 1.125 & 1.164 & 1.04 & 0.001 \\
\quad & pyramid & $1.6\times\sigma_s^\ast$ & \textbf{1.311} & \textbf{1.417} & \textbf{1.13} & 0.006 \\
\quad & pyramid & $2.0\times\sigma_s$      & 1.286 & 1.561 & 1.11 & 0.003 \\

\bottomrule
\end{tabular}

\vspace{0.4em}
\footnotesize
$^\ast$ Selected visualization run from the pooled pyramid sweep.
Diagnostics are used to screen for collapse and latent-space usage; final
selection also uses global visual topology and mapped-back spatial
interpretability. For MHD, the fixed-box rows provide the large-occlusion
control; all fixed-box runs reach near-complete hinge saturation.
The $2.0\times\sigma_s$ pyramid MHD endpoint similarly saturates, so the
$1.2\times\sigma_s$ run is used for visualization.
NGC runs use gradient accumulation with factor~2 (because the limited
number of available targets per field makes single-sample batches noisy)
and a mask hard cap of 35~px; the selected run is $1.6\times\sigma_s$.
\end{table}

\section{Plain ConvNeXt control without scale hierarchy}
\label{app:convnext_control}

To test whether the latent organization recovered by ScaleAware-JEPA can be
explained by generic masked-image ConvNeXt capacity alone, we include a plain
dense ConvNeXt control that removes the CDD pyramid and all scale-aware
processing. The control receives only two input channels: the masked scalar
field and a binary mask-indicator map. It uses the same overall ConvNeXt
design language as the main model---depth $4$, hidden width $64$, latent
width $32$, kernel size $7$, GRN, LayerNorm, and reflect padding---but has
no CDD channels, no per-scale adapters, and no scale-aware fusion. In other
words, it is a matched JEPA backbone that must infer all multiscale
organization directly from raw masked pixel intensities.

We evaluated this control on MHD across box footprints of
$7$, $11$, $15$, and $19$\,px.
All runs produce nearly identical diagnostics (Table~\ref{tab:mhd_convnext_box_control}),
with saturation behavior observed across all tested footprints. Varying the
box footprint therefore yields no meaningful change in latent geometry.

\begin{figure}[H]
\centering
\includegraphics[width=\textwidth]{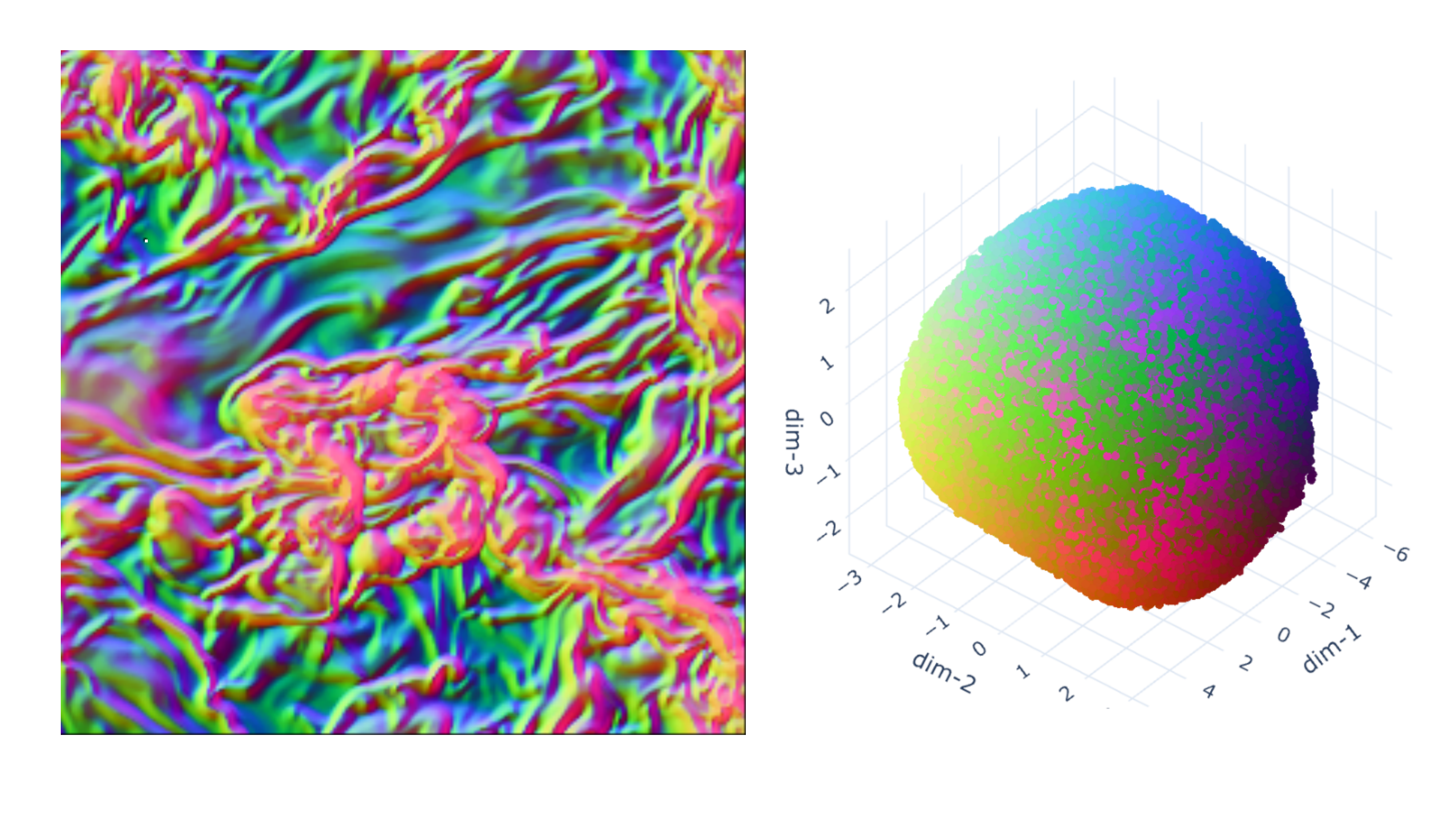}
\caption{\textbf{Plain ConvNeXt control versus the scale-aware encoder.}
Comparison on a matched random-mask example. The dense ConvNeXt control
receives only the masked scalar field and a binary mask-indicator channel,
whereas ScaleAware-JEPA receives pixel-registered multiscale CDD components.
In the control runs, varying the box footprint from $7$ to $19$\,px does
not produce a meaningful change in sampled embedding usage and the hinge ratio
remains saturated, indicating that generic masked-image ConvNeXt capacity alone
does not recover the scale-dependent latent organization seen in the main
model, and that small-scale structures are over-sharpened relative to the
scale-aware encoder output.}
\label{fig:convnext_comparison}
\end{figure}

This behavior contrasts with the scale-aware MHD sweep, where changing the
scale-tied masking footprint produces a structured progression from thin,
under-constrained embeddings to richer middle-regime geometry before entering
the large-occlusion saturated regime. In the plain ConvNeXt control, by
contrast, the hinge simply saturates without revealing comparable scale
dependence, and small-scale structures appear over-sharpened
(Figure~\ref{fig:convnext_comparison}). This suggests that the multiscale
organization recovered by the main model is not explained by generic masked-image ConvNeXt capacity alone,
but depends on providing the encoder and masking operator with explicit
physical scale coordinates.

\begin{table}[H]
\centering
\caption{\textbf{ConvNeXt ablation MHD diagnostics.}
Plain ConvNeXt image encoder without CDD frontend, evaluated across fixed-box
mask footprints. All runs use pooled standard-deviation hinge regularization and
produce near-identical diagnostics with saturated hinge ratios, indicating that
without CDD scale channels the encoder cannot use the scale-informed masking
hierarchy. The $7$\,px run is selected for visualization.}
\label{tab:mhd_convnext_box_control}

\setlength{\tabcolsep}{5pt}
\renewcommand{\arraystretch}{1.05}

\begin{tabular}{r r r r}
\toprule
Mask footprint & Target rank & Pred. rank & Hinge ratio \\
\midrule
$7$ px$^\ast$  & \textbf{3.866} & \textbf{5.752} & 1.000 \\
$11$ px        & 3.914 & 5.771 & 1.000 \\
$15$ px        & 3.906 & 5.920 & 1.000 \\
$19$ px        & 3.906 & 5.920 & 1.000 \\
\bottomrule
\end{tabular}
\end{table}

\FloatBarrier

\end{document}